\documentclass[10pt,twocolumn,letterpaper]{article}

\usepackage[pagenumbers]{cvpr} % To force page numbers, e.g. for an arXiv version
\usepackage{titletoc}
\usepackage[dvipsnames]{xcolor}

\usepackage{graphicx}
\usepackage{amsmath}
\usepackage{amssymb}
\usepackage{booktabs}
\usepackage{comment}
\usepackage{multirow}

\usepackage{pifont}
\usepackage{soul}
\usepackage{colortbl}
\usepackage{comment}
\usepackage{wrapfig, lipsum, booktabs}
\usepackage{makecell}
\usepackage{enumitem}
\usepackage{amsmath}
\usepackage{bm}
\usepackage{nicematrix}
\usepackage{adjustbox}

\definecolor{cvprblue}{rgb}{0.21,0.49,0.74}
\usepackage[pagebackref,breaklinks,colorlinks,citecolor=cvprblue]{hyperref}

\title{MicroCinema: A Divide-and-Conquer Approach for Text-to-Video Generation}

\author{Yanhui Wang\textsuperscript{1,2$\ast\dagger$}, Jianmin Bao\textsuperscript{2$\ast$}, Wenming Weng\textsuperscript{1}, Ruoyu Feng\textsuperscript{1}, Dacheng Yin\textsuperscript{1}, Tao Yang\textsuperscript{3}, \\
Jingxu Zhang\textsuperscript{1},
Qi Dai\textsuperscript{2},
Zhiyuan Zhao\textsuperscript{2}, Chunyu Wang\textsuperscript{2}, Kai Qiu\textsuperscript{2}, Yuhui Yuan\textsuperscript{2},\\Chuanxin Tang\textsuperscript{2},
Xiaoyan Sun\textsuperscript{1}, Chong Luo\textsuperscript{1,2$\ddagger$}, Baining Guo\textsuperscript{2}\\
{\textsuperscript{1}University of Science and Technology of China
~\textsuperscript{2}Microsoft Research Asia~\textsuperscript{3}Xi'an Jiaotong University}\\
{\textsuperscript{$\ast$}Equal Contribution
~~\textsuperscript{$\dagger$}This work was done during the internship at MSRA~~\textsuperscript{$\ddagger$}Corresponding author.}}

\begin{document}
% \maketitle
\twocolumn[{%
\renewcommand\twocolumn[1][]{#1}%
\maketitle
\begin{center}
    \centering
    \captionsetup{type=figure}
    \vspace{-3mm}
    \includegraphics[width=1.0\linewidth]{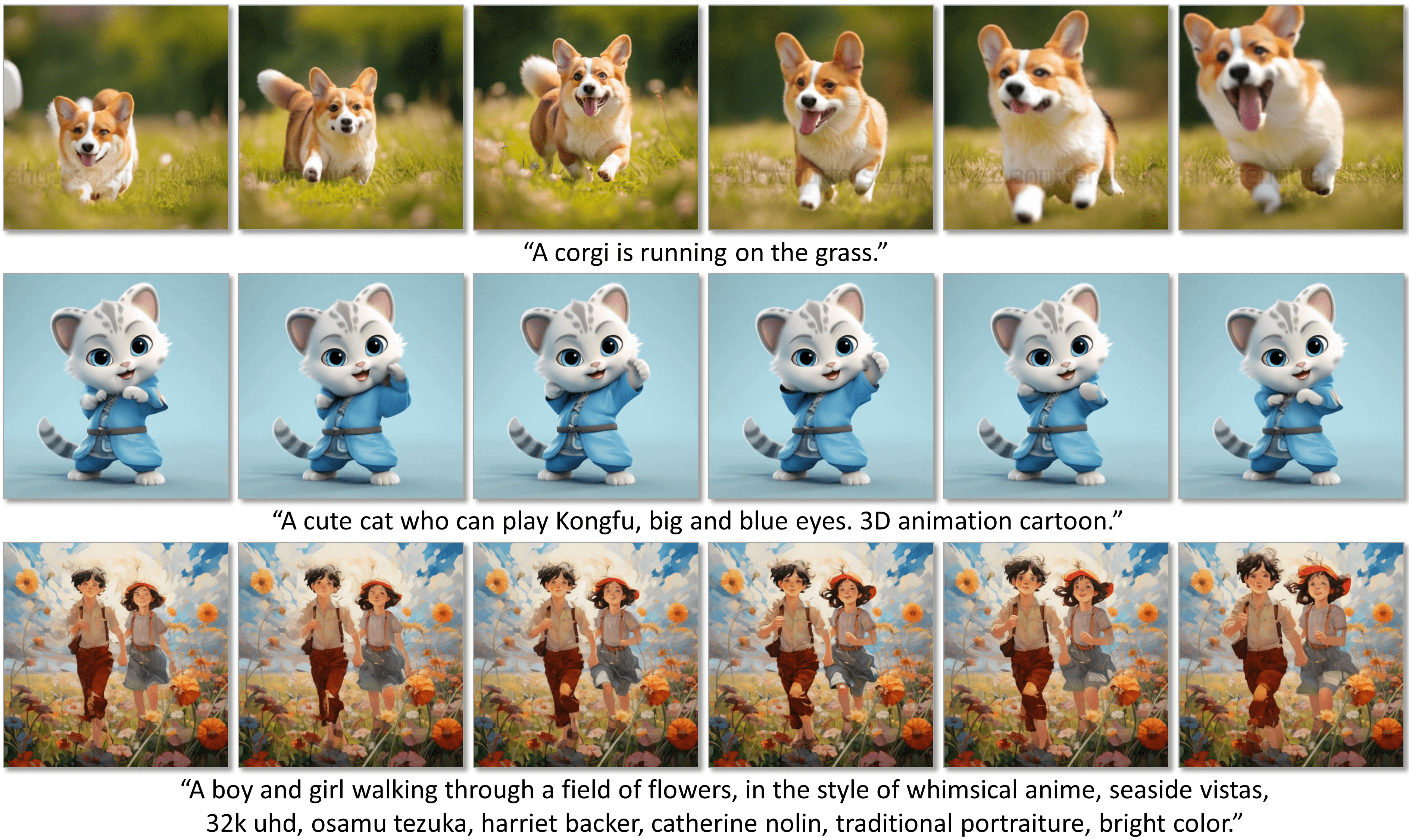}

    \captionof{figure}{Sample videos produced by MicroCinema, our proposed text-to-video generation system. They showcase MicroCinema's ability to create coherent and high-quality videos, with precise motion aligned with text prompts. Image reference generated by Midjourney.}
    \label{fig: teaser}
\end{center}%
}]
\begin{abstract}
\vspace{-1.0em}
We present MicroCinema, a straightforward yet effective framework for high-quality and coherent text-to-video generation. Unlike existing approaches that align text prompts with video directly, MicroCinema introduces a Divide-and-Conquer strategy which divides the text-to-video into a two-stage process: text-to-image generation and image\&text-to-video generation. This strategy offers two significant advantages. a) It allows us to take full advantage of the recent advances in text-to-image models, such as Stable Diffusion, Midjourney, and DALLE, to generate photorealistic and highly detailed images. b) Leveraging the generated image, the model can allocate less focus to fine-grained appearance details, prioritizing the efficient learning of motion dynamics. 
To implement this strategy effectively, we introduce two core designs. First, we propose the Appearance Injection Network, enhancing the preservation of the appearance of the given image. Second, we introduce the Appearance Noise Prior, a novel mechanism aimed at maintaining the capabilities of pre-trained 2D diffusion models. These design elements empower MicroCinema to generate high-quality videos with precise motion, guided by the provided text prompts. Extensive experiments demonstrate the superiority of the proposed framework. Concretely, MicroCinema achieves \textbf{SOTA} zero-shot FVD of \textbf{342.86} on UCF-101 and \textbf{377.40} on MSR-VTT. See \href{https://wangyanhui666.github.io/MicroCinema.github.io/}{project page} for video samples.
\end{abstract}    
\section{Introduction}
\label{sec:intro}

Diffusion models~\cite{ho2020denoising, song2020score} have achieved remarkable success in text-to-image generation, such as DALL-E~\cite{ramesh2021zero}, Stable Diffusion~\cite{rombach2022high}, Imagen~\cite{saharia2022photorealistic}, among others. They can generate unseen image content based on novel text concepts, showcasing impressive capabilities for image content generation and manipulation. Consequently, researchers have sought to extend the success of diffusion models to text-to-video generation.

One prevalent strategy involves training large-scale text-to-video diffusion models directly~\cite{ho2022video, ho2022imagen, villegas2022phenaki, singer2022make}. These models employ cascade spatiotemporal diffusion models to learn from text and video pairs. While capable of producing high-quality videos, they pose challenges due to substantial GPU resource requirements and the need for extensive training data. Recently, some works~\cite{blattmann2023align, ge2023preserve} have presented a cost-effective strategy. These methods entail the insertion of temporal layers into a text-to-image model, followed by fine-tuning on paired text and video data to create a text-to-video model. However, videos generated using this approach may encounter issues related to appearance and temporal coherence. We argue that maintaining appearance and temporal coherence is crucial for effective video generation.

In this paper, we present a novel approach, named MicroCinema, which employs a divide-and-conquer strategy to address appearance and temporal coherence challenges in video generation. The model features a two-stage generation pipeline. In the first stage, we generate a center frame, which serves as the foundation for subsequent video clip generation based on the input text. This design offers the flexibility to utilize any existing text-to-image generator for the initial stage, allowing users to incorporate their own images to establish the desired scene.

The second stage, known as image\&text-to-video, concentrates on motion modeling. To achieve this, we leverage the open-source text-to-image generation model called Stable Diffusion (SD)~\cite{rombach2022high} and inject temporal layers into it to obtain a three-dimensional (3D) network structure. The SD model has been trained on the filtered large-scale LAION dataset~\cite{schuhmann2022laion}. Its strong performance in generating high-quality images demonstrates its ability to capture spatial information within visual signals. To further enhance the model's ability to capture motion, we propose two core designs for the image\&text-to-video model. 

First, we introduce an Appearance Injection Network to inject the given image as a condition to guide the video generation. Concretely, it shares the structure of the encoder and middle part of the 3D U-Net and feeds the learned feature into the main branch via dense injection in a multi-layer manner. The dense injection operation better injects the appearance into the main branch, thus releasing the model from appearance modeling and encouraging the model dedicated to motion modeling. Second, we propose an appearance-aware noise strategy to preserve the pre-trained capability of the SD model by modifying the i.i.d. noise in the diffusion process. Specifically, we add an appropriate amount of center frame to the i.i.d. noise without altering the overall diffusion training and inference process. This appearance-aware noise provides an intuitive cue to the model to generate a video whose appearance is similar to the given center frame, thereby unleashing its motion modeling capabilities.

Equipped with these designs, our framework can generate appearance-preserving and coherent videos with a given image and text. Extensive experiments demonstrate the superiority of MicroCinema. We achieve a \textbf{state-of-the-art} zero-shot FVD of \textbf{342.86} on UCF101~\cite{soomro2012ucf101} and \textbf{377.40} on MSR-VTT~\cite{xu2016msr} when training on the public WebVid-10M~\cite{bain2021frozen} dataset.

%\textcolor{red}{todo: explain our contributions}

In summary, our contributions are presented as follows:
\begin{itemize}
    \item We introduce an innovative two-stage text-to-video generation pipeline that capitalizes on a key-frame image generated by any off-the-shelf text-to-image generator in the initial stage. Subsequently, both the generated key-frame image and text serve as inputs for the video generation process in the second stage.
    \item We propose an Appearance Injection Network structure to encourage the 3D model to focus on motion modeling during the image\&text-to-video generation process. 
    \item We introduce an effective and distinctive Appearance Noise Prior tailored for fine-tuning text-to-image diffusion models. This modification significantly elevates the quality of video generation.
    %\item  We conducted systematic studies on the WebVid10M dataset, carefully identifying effective rules for data cleaning. This effort has led to the creation of the WebVid10M-Cleaned dataset, which has proven to be more suitable for training video generation models.  
    \item In-depth quantitative and qualitative results are presented to validate the video generation capability of our proposed MicroCinema.
\end{itemize}
\section{Related Work}
\label{sec:related}
The task of video generation involves addressing two fundamental challenges: image generation and motion modeling. Various approaches have been employed for image generation, including Generative Adversarial Networks (GANs)~\cite{goodfellow2020generative,salimans2016improved,arjovsky2017towards,arjovsky2017wasserstein}, Variational autoencoder (VAE)~\cite{hinton2006reducing,razavi2019generating} and flow-based methods~\cite{dinh2015nice}. Recently, the state-of-the-art methods are built on top of diffusion models such as DALLE-2~\cite{ramesh2022hierarchical}, Stable Diffusion~\cite{rombach2022high}, GLIDE~\cite{nichol2021glide} and Imagen~\cite{saharia2022photorealistic}, which achieved impressive results. Extending these models for video generation is a natural progression, though it necessitates non-trivial modifications.

\noindent\textbf{Text-to-Video Models.} Image diffusion models adopt 2D U-Net with few exceptions~\cite{peebles2022scalable}. To generate temporally smooth videos, temporal convolution (conv) or attention layers are also introduced. Notably, in Align-your-latents~\cite{blattmann2023align}, 3D conv layers are interleaved with the existing spatial layers to align individual frames in a temporally consistent manner. This factorized space-time design has become the de facto standard and has been used in VDM~\cite{ho2022video}, Imagen Video~\cite{ho2022imagen}, and CogVideo~\cite{hong2022cogvideo}. Besides, it creates a concrete partition between the pre-trained two-dimensional (2D) conv layers and the newly initialized temporal conv layers, allowing us to train the temporal convolutions from scratch while retaining the previously learned knowledge in the spatial convolutions’ weights. More recent work Latent-Shift~\cite{an2023latent} introduces no additional parameters but shifts channels of spatial feature maps along the temporal dimension, enabling the model to learn temporal coherence.   Many approaches rely on temporal layers to implicitly learn motions from paired text and videos~\cite{singer2022make,ho2022video,ho2022imagen,hong2022cogvideo,blattmann2023align}.  The generated motions, however, still lack satisfactory global coherence and fail to faithfully capture the essential movement patterns of the target subjects.

\noindent\textbf{Leveraging Prior for Text-to-Video Diffusion Models.}  Generating natural motions poses a significant challenge in video generation. Many attempts are focused on leveraging prior into the text-to-video generation process. ControlVideo~\cite{zhang2023controlvideo} directly utilizes ground truth motions, represented as depth maps or edge maps, as conditions for video diffusion models, demonstrating the importance of motion in video generation. GD-VDM~\cite{lapid2023gd} involves a two-phase generation process leveraging generating depth videos followed by a novel diffusion Vid2Vid model that generates a coherent real-world video in the autonomous driving scenario. However, it is not clear whether it can be applied to general scenes due to the lack of depth training data. 
Make-Your-Video~\cite{xing2023make} utilizes a standalone depth estimator to extract depth from a driving video, bypassing the need for depth generation, to generate new videos. In Leo~\cite{wang2023leo}, a motion diffusion model is trained to generate a sequence of motion latents, fed to a decoder network to recover the optical flows to animate the input image. Meanwhile, other methods involve linear displacement of codes in latent space~\cite{wang2022latent}, noise correlation~\cite{khachatryan2023text2video}, and generating textual descriptions for motion~\cite{hong2023large}, serving as conditions for video generation models. More recent work PYoCo~\cite{ge2023preserve} proposes the video diffusion noise prior for a diffusion model and cost-effectively fine-tuning the text-to-image model. 

Our proposed framework differs significantly from existing methods by employing a Divide-and-Conquer strategy. In our approach, we first generate images and subsequently capture motion dynamics along the temporal dimension. We also notice that a previous method Make-A-Video~\cite{singer2022make} has adopted a similar approach. However, our method introduces a novel model network design and incorporates an appearance-noise prior. This innovation ensures the generated video not only maintains the appearance established in the initial stage but also demonstrates superior motion modeling capabilities, a feature notably absent in Make-A-Video and concurrently related methods~\cite{li2023videogen, zhang2023i2vgen}.
\section{MicroCinema}
\label{sec:method}

\begin{figure*}[ht]
  \centering
  \includegraphics[width=1.0\linewidth]{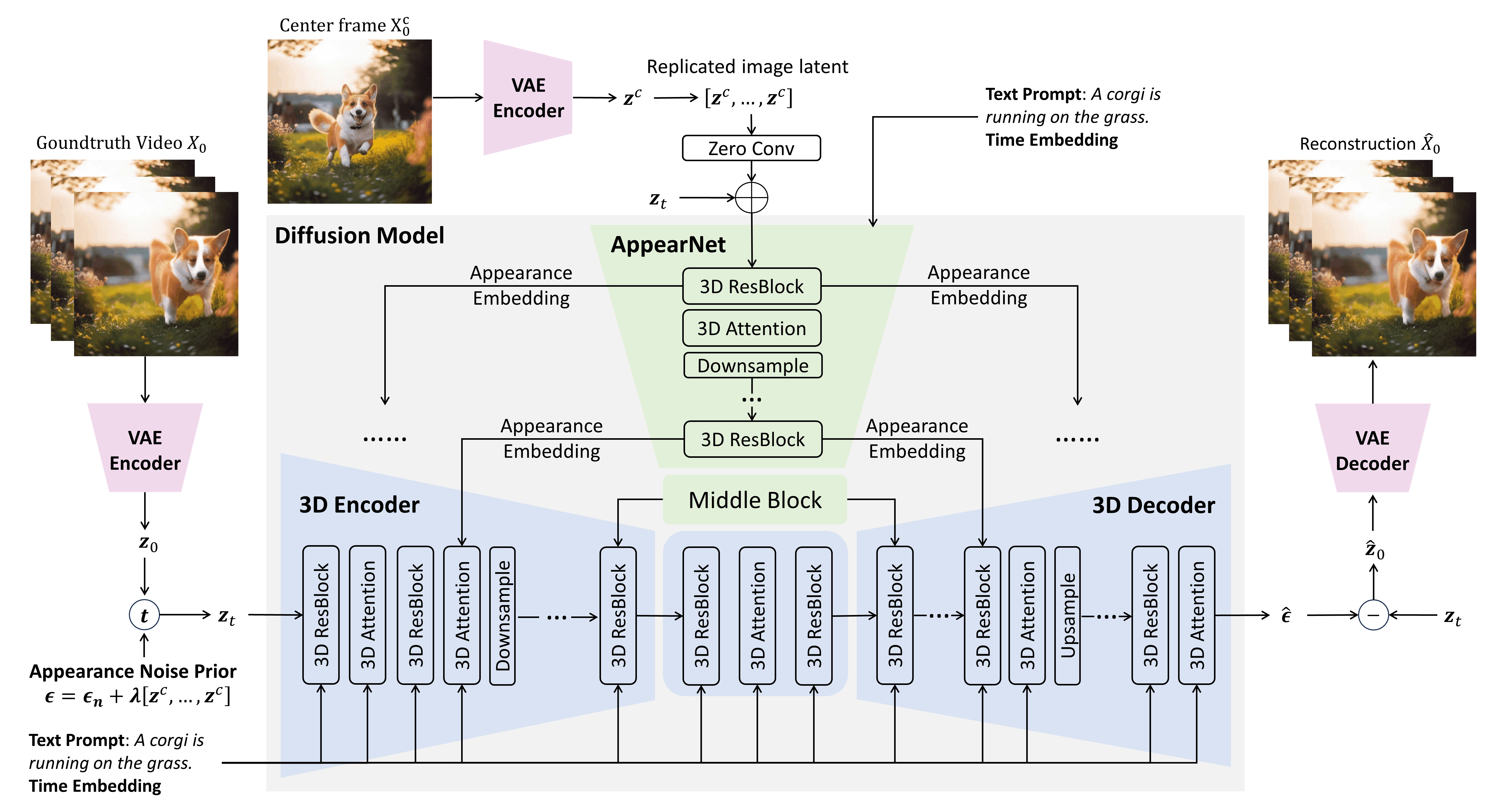}
  \vspace{-8mm}
  \caption{Overall architecture of our proposed diffusion-based image\&text-to-video model in MicroCinema. The proposed AppearNet aims to provide appearance information for video generation.}

    \vspace{-5mm}
  \label{fig:overall_architecture}
\end{figure*}

\subsection{Overview}
% inflating 2d to 3d and temporal attn
%Our methodology employs a diffusion-based generative model.

% 1. Two-stage architecture. 
Our approach decomposes the text-to-video generation process into two distinct stages. Initially, we employ prevalent off-the-shelf text-to-image generation techniques to produce a key frame. Subsequently, both the key frame, acting as the center frame, and the text prompts are used as input to the image\&text-to-video model to generate videos. We argue that the image\&text-to-video model in a two-stage framework exhibits the potential for yielding more natural videos compared to the single-stage text-to-video model. This argument rests on the premise that by incorporating the center frame as a condition, our approach mitigates the model's burden in learning complicated appearance.

In the image\&text-to-video generation stage, we adopt a cascaded approach to produce high-quality videos. First, we use a base image\&text-to-video model to generate low frame rate videos from given image and text. Then, an adapted temporal interpolation model, derived from the base model, is employed to augment the frame rate. Finally, an off-the-shelf spatial super-resolution model is incorporated to render high-definition videos. This paper focuses on explaining the base model design and detailing its adaptation into the temporal interpolation model.
% Then, we apply the off-the-shelf models~\cite{} for spatial super-resolution(SSR) to increase the resolution to $1280\times1280$. 

\noindent\textbf{Base image\&text-to-video model.} \cref{fig:overall_architecture} illustrates the overall architecture of the base image\&text-to-video model in MicroCinema. This model is extended from the widely recognized Stable Diffusion (SD) model \cite{rombach2022high}. 
%\subsection{Image\&Text-to-Video Diffusion Models}\label{sec:3.2}
% MotionNet and dense connection
%we first introduce the network structure of the base image\&text-to-video model.
Following previous attempts~\cite{singer2022make, blattmann2023align}, we first extend the 2D U-Net into a 3D structure. We first enhance the original model by adding a 1D temporal convolution (conv) layer following each 2D spatial conv layer, enhancing its ability to handle temporal alignments. Additionally, we introduce a 1D temporal attention layer after every 2D spatial attention layer. These attention layers effectively capture long-range temporal correspondence, complementing the functionality of the 1D conv layers. To protect the strong capability of SD, we zero-initialize all the convolution and attention temporal layers and add a skip connection to it. Based on these modifications, we obtain a 3D model that can handle text-to-video generation.
The base image\&text-to-video model showcases two crucial innovations: the AppearNet and the appearance noise prior. Both are designed to incorporate appearance information from the key frame. A detailed explanation of these technical advancements will be provided in \cref{sec:3.2} and \cref{sec:3.3}.

\noindent\textbf{Temporal interpolation model.} Our base model generates videos at a resolution of $320 \times 320$ pixels with a frame rate of 2 frames per second (fps). To enhance temporal quality, we train a temporal interpolation model designed for four-fold temporal super-resolution (TSR). This TSR model mirrors the architecture of the base model with slight modifications. The base model employs only one conditional image (the center frame) while there are two conditional images (the start and end frames) in the TSR model. Accordingly, we alter the input of the AppearNet, shifting from duplicating the center frame to utilizing the interpolated latent representations of the given first and last frames. Leveraging this model consecutively on adjacent frames from previous steps boosts the frame rate from 2 fps to 32 fps.

\begin{figure}[tb!]
  \centering
  \includegraphics[width=0.95\linewidth]{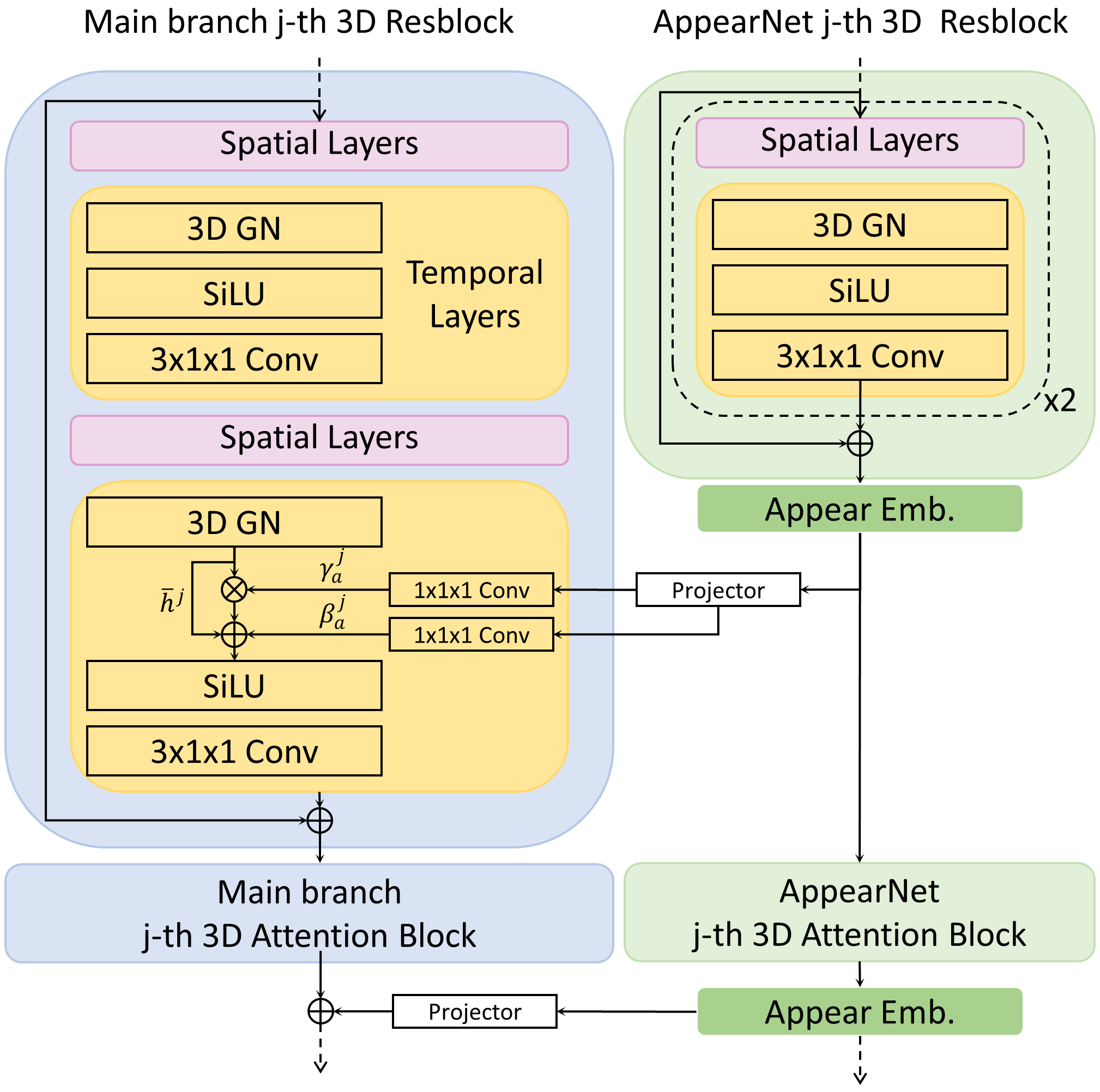}
  \caption{AppearNet injects multi-scale features into the main branch to perform a dense fusion.}
  \vspace{-1.0em}
  \label{fig:inject}
\end{figure}

\subsection{Appearance Injection Network} \label{sec:3.2}
To enhance the model's capability in handling reference center frame, we introduce the Appearance Injection Network, abbreviated as AppearNet, to the 3D network as depicted in \cref{fig:overall_architecture}. Inspired by ControlNet~\cite{blattmann2023align}, we let AppearNet inherit the encoder and the middle part of the backbone network. Let $N$ be the frame length of the output video. Then the center frame $\bm{z^c}$ is replicated for $N$ times to create an image sequence, denoted as $[\bm{z^c}, \bm{z^c}, \hdots, \bm{z^c}]$. It is used as input to the AppearNet to offer a robust appearance cue for generating output video frames.

We apply a multi-scale and dense fusion mechanism to seamlessly integrate the outputs of the AppearNet into the main branch. The multi-scale output of AppearNet is injected into both the encoder and the decoder of the main branch at the corresponding scales. In addition to the commonly used additive operation, we introduce an effective strategy of de-normalization~\cite{park2019semantic} to inject the feature into the corresponding normalization layer of the main branch. As shown in \cref{fig:inject}, at the $j$-th feature level, let $\bm{h}_{m}^j$ denote the activation map in the main branch. Before integration, we perform 3D Group Normalization~\cite{wu2018group} on $\bm{h}_{m}^j$: 
\vspace{-0.5em}
\begin{equation}
\vspace{-0.5em}
\label{eqn:IN}
\bar{\bm{h}}^j = \frac{\bm{h}_{m}^j - \bm{\mu}^j}{\bm{\sigma}^j}.
\end{equation}
Here $\bm{\mu}^j$ and $\bm{\sigma}^j$ are the means and standard deviations of $\bm{h}_{m}^j$'s group-wise activations. 
For AppearNet feature integration, let $\bm{f}_{a}^j$ be the AppearNet embedding on this feature level, we compute the output activation $\bm{o}^j$ by denormalizing the normalized $\bar{\bm{h}}^j$ according to $\bm{f}_{a}^j$, formulated as
\vspace{-0.5em}
\begin{equation}
\vspace{-0.5em}
\label{eqn:SPADE}
\bm{o}^j = (\gamma^j_{a} + 1) \otimes \bar{\bm{h}}^j + \beta_{a}^j,
\end{equation}
where $\gamma^j_{a}$ and $\beta^j_{a}$ are obtained by convolving from the feature map $\bm{f}_{a}^j$.
The computed $\gamma^j_{a}$ and $\beta^j_{a}$ are multiplied and added to $\bar{\bm{h}}^j$ in an element-wise manner. Equipped with this design, our entire structure could better maintain the appearance from a given center frame while possessing the ability to generate videos based on text and image conditions.

\subsection{Appearance Noise Prior} \label{sec:3.3}

Fine-tuning from a text-to-image model proves to be a cost-effective approach for acquiring a video generation model. However, this process presents challenges due to the transition of the output space from images to videos. In the context of a typical T2I diffusion model, it tends to generate appearance-irrelevant images from a sequence of independent noise (sampled from $\mathcal{N}(\bm 0, \bm I)$). In video generation, a sequence of independent noise should ideally yield a video with a coherent appearance. Therefore, the fine-tuning process may potentially compromise the capability of the original 2D T2I model. Our focus lies in preserving the effectiveness of the original 2D T2I model during the fine-tuning process for the image\&text-to-video model.  
    
For our proposed image\&text-to-video model, the model should expand the given center image to a sequence of frames, which have a similar appearance to the center frame. Consequently, the output video is predominantly determined by the center frame rather than the sampled noise in the original diffusion process. To address this, we modify the noise distribution to align with the appearance of the given center frame. Leveraging the denoising property of the diffusion model, we introduce Appearance Noise Prior by adding an appropriate amount of the center frame into the noise, in order to generate appearance-conditioned frames.

Let $\bm \epsilon = [\bm \epsilon^{1}, \bm \epsilon^{2}, \hdots, \bm \epsilon^{N}]$ denote the noise corresponding to a video clip with $N$ frames, $\bm \epsilon^{i}$ represents the noise added to the $i^{th}$ frame. $\bm{z}^c$ is the latent tensor of center frame,  $\bm \epsilon_{n}^{i}$ is the randomly sampled noise from $\mathcal{N}(\bm 0, \bm I)$. The training noise for our model is defined as:
\vspace{-3mm}
\begin{align}\label{eq:appearance_noise_prior} 
    \bm \epsilon^{i} = \lambda \bm{z}^c + \bm \epsilon_{n}^{i}, 
\end{align}
where $\lambda$ is the coefficient that controls the amount of the center frame.

Consequently, the diffusion process of our model can be expressed in the following form, the t-step noisy input of the diffusion model is:
\vspace{-3mm}
\begin{equation}
\bm{z}_t = \sqrt{\bar{\alpha}_t} \bm{z}_0 + \sqrt{1 - \bar{\alpha}_t} \bm \epsilon
\end{equation}
where $\bm{z}_0$ is the latent tensors of an input video and $\bar{\alpha}_t$ is the same as defined in DDPM~\cite{ho2020denoising}.

For training, we adhere to the stable diffusion training setting and use noise prediction with the following loss function:

\begin{equation}
\mathcal{L}_{\bm{\theta}} = \mathbb{E}_{q_{t}(\bm{z}_0, \bm{z}_t)} \left[ \left \| \bm{f}_{\bm{\theta}}(\bm{z}_t, t, \bm{z}^c, \bm{c})-\bm{\epsilon} \right \|^2 \right],
\end{equation}
where $t$ is the time step, $\bm{z}^c$ is the reference image input, $\bm{c}$ is the text input, $\bm z_0$, $\bm z_t$ are the ground-truth video and noisy input, $\bm{f}_{\bm{\theta}}(\bm{z}_t, t, \bm{z}^c, \bm{c})$ represents the output of the model., respectively. Our appearance noise prior employs the same inference strategy as previous methods, differing only in the initiation of noise, which aligns with our formulation. This consistency allows for the direct application of existing ODE sample algorithms. For a thorough understanding of the proofs, please refer to the supplementary materials.

\section{Experiments}
\label{sec:experiments}

\noindent\textbf{Datasets.}
MicroCinema is trained using the public WebVid-10M dataset \cite{bain2021frozen}, comprising ten million video-text pairs. This dataset exhibits a wide spectrum of video motions, ranging from near-static sequences to those with frequent and abrupt scene changes. Text captions are automatically sourced from alt text, resulting in some noise. Therefore, we perform a filtering process which excludes video-text pairs with a low CLIP score or with excessively high or low motions.

\begin{table}[t]
\centering
\caption{\small Comparison on the zero-shot text-to-video generation performance on UCF-101\cite{soomro2012ucf101} and MSR-VTT\cite{xu2016msr}}
\label{tab:zeroshot}
\renewcommand{\arraystretch}{1.0}
\begin{small}
\begin{tabular}{@{}l|cc|cc@{}}
\toprule
Methods & \multicolumn{2}{c|}{UCF-101\cite{soomro2012ucf101}} & \multicolumn{2}{c}{MSR-VTT\cite{xu2016msr}} \\ 
              & FVD $\downarrow$ & IS $\uparrow$ & FVD $\downarrow$ & CLIPSIM $\uparrow$ \\
\midrule
\multicolumn{5}{c}{\textit{Using WebVid-10M and additional data for training}}\\
Make-A-Video \cite{singer2022make}             & 367.23 & 33.00  & -       & 0.3049 \\
VideoFactory \cite{wang2023videofactory}       & 410.00 & -      & -       & 0.3005 \\
ModelScope \cite{wang2023modelscope}           & 410.00 & -      & 550.00  & 0.2930 \\
Lavie \cite{wang2023lavie}                     & 526.30 & -      & -       & 0.2949 \\
VidRD \cite{gu2023reuse}                       & 363.19 & 39.37  & -       & -      \\
PYoCo \cite{ge2023preserve}                    & 355.19 & \textbf{47.76}  & -       & \textbf{0.3204} \\
\midrule
\multicolumn{5}{c}{\textit{Using WebVid-10M only for training}}\\
LVDM \cite{he2023latent}                       & 641.80 & -      & 742.00  & 0.2381 \\
CogVideo \cite{hong2022cogvideo}               & 701.59 & 25.27  & 1294    & 0.2631 \\
MagicVideo \cite{zhou2022magicvideo}           & 699.00 & -      & 998.00  & -      \\
Video LDM \cite{blattmann2023align}            & 550.61 & 33.45  & -       & 0.2929 \\
VideoComposer \cite{wang2023videocomposer}     & -      & -      & 580     & 0.2932 \\
VideoFusion \cite{luo2023videofusion}          & 639.90 & 17.49  & 581.00  & 0.2795 \\
SimDA \cite{xing2023simda}                     & -      & -      & 456.00  & 0.2945 \\
Show-1 \cite{zhang2023show}                    & 394.46 & 35.42  & 538.00  & 0.3072 \\
%\midrule
%& MicroCinema Base(Ours)                         & 336.38 & 32.07  & 377.40  & 0.2967 \\
MicroCinema (Ours)                    & \textbf{342.86} & 37.46  & \textbf{377.40}  & 0.2967    \\
\bottomrule
\end{tabular}
  \vspace{-1.5em}
\end{small}
\end{table}

\begin{figure*}[ht]
  \centering
  \includegraphics[width=0.95\linewidth]{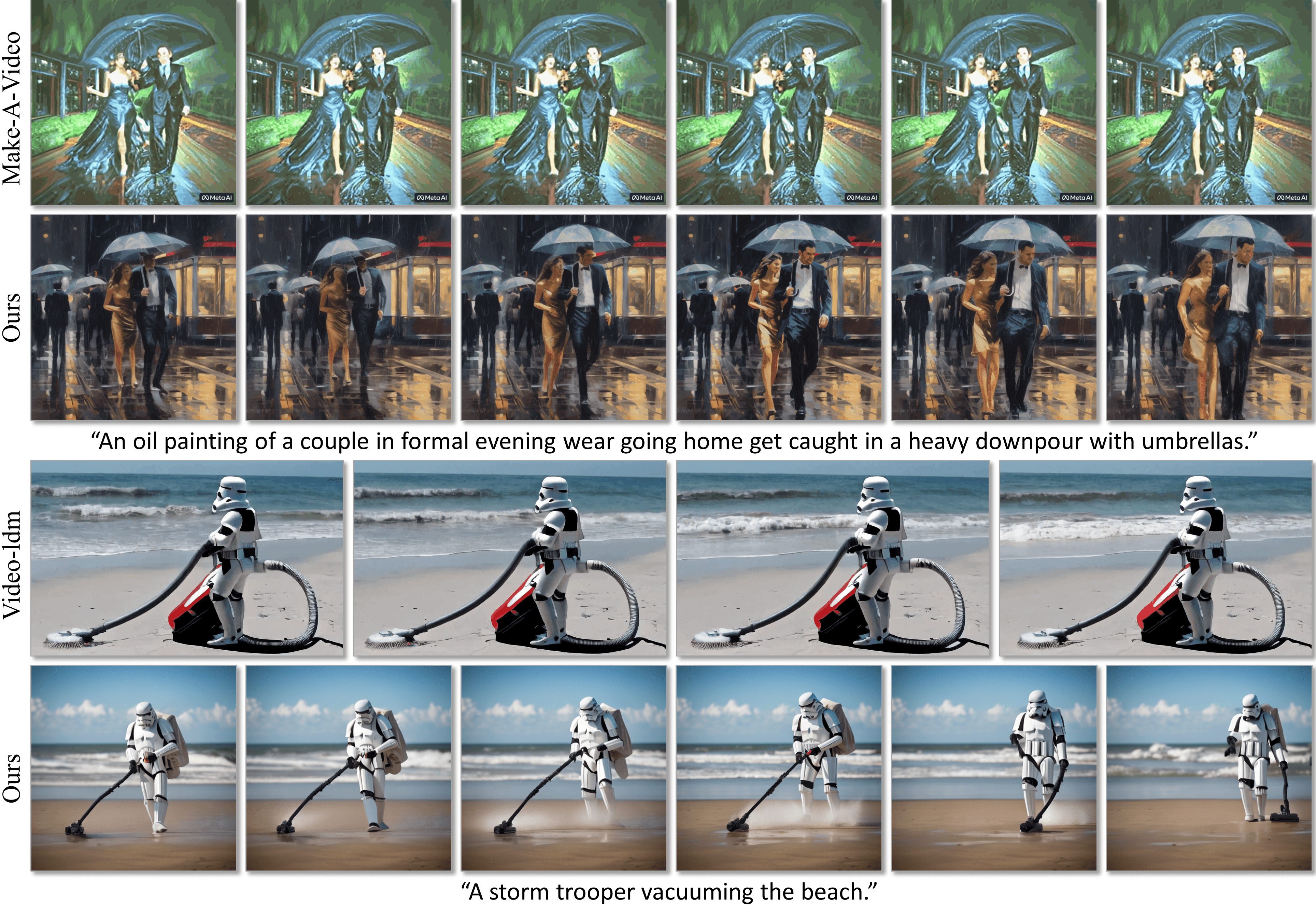}
    \vspace{-0.5em}
  \caption{Comparison with Make-A-Video and Video LDM. Reference images generated by DALL-E 2 (top) and Midjourney (bottom). The generated videos from our model shows a clear and coherent motion.}
  \label{fig:compare_vis}
  \vspace{-1.5em}
\end{figure*}

\noindent\textbf{Evaluation metrics.}
The quantitative evaluations are conducted on UCF-101 \cite{soomro2012ucf101} and MSR-VTT \cite{xu2016msr} benchmark datasets under the zero-shot setting.
On UCF-101, Frechet Video Distance (FVD) \cite{unterthiner2019fvd} and Inception Score (IS) \cite{salimans2016improved} are reported to validate the temporal consistency, where 10K or 2K video clips are generated using a sentence template of the category names.
On MSR-VTT, Frechet Inception Distance (FID) \cite{heusel2017gans} and CLIPSIM \cite{wu2021godiva,singer2022make} are provided to assess the quality of generated frames and the semantic correspondence, where CLIPSIM is computed by averaging the cosine similarity of CLIP embeddings \cite{radford2021learning} between generated frames and captions.
We utilized captions from the MSR-VTT validation set, comprising 2.9K entries, to generate the video clips.
The condition images are generated with SDXL model \cite{rombach2022high} on all evaluations unless otherwise specified.

\subsection{Comparison with State-of-the-Arts}

\noindent\textbf{Implementation details.} 
%Our approach follows a two-stage methodology. In the first stage, we can utilize various models for text-to-image generation, including SD2.1, SDXL, DALL·E, Midjourney, and others. 
MicroCinema generates video from text in a two-stage process. In the first stage, we employ a SOTA T2I model SDXL \cite{podell2023sdxl} to generate an image according to the text. Then in the second stage, the image\&text-to-video generation model is built upon the pre-trained weights of Stable Diffusion 2.1. Temporal layer is zero initialized. During training, the learning rate for the temporal modules is set to 2e-5, while the learning rate for the spatial model is 10 times smaller than that of the temporal modules. The output of the image\&text-to-video model yields a video clip with a spatial resolution of 320x320, consisting of 9 frames at a rate of 2fps. The model is trained on the filtered WebVid dataset for one epoch, employing the same diffusion noise schedule as SD2.1.

\noindent\textbf{Quantitative evaluation.} 
We evaluate zero-shot text-to-video generation performance on both UCF101 and MSR-VTT. In the case of UCF101, we produce 10K samples using simple clip captions. For MSR-VTT, we generate 2.9K samples using the captions provided within the MSR-VTT dataset. \cref{tab:zeroshot} presents a quantitative comparison between MicroCinema and alternative text-to-video models. These models are categorized into two groups based on whether they leverage additional data beyond WebVid-10M. 
As data is of paramount importance to the training of video generation model, we can observe that the methods in the first group (with additional data) achieve superior overall performance compared to those in the second group. 
Remarkably, despite being exclusively trained on the WebVid-10M dataset, our proposed MicroCinema, with its innovative design, achieves the most outstanding performance among all methods on both datasets. It achieves the lowest FVD values of 342.86 on UCF101 and 377.40 on MSR-VTT. Notably, MicroCinema surpasses methods employing additional data and notably outperforms those relying solely on the WebVid-10M dataset by a considerable margin.

\noindent\textbf{Qualitative evaluation}
\cref{fig:compare_vis} compares the video clips generated by MicroCinema and two other methods, known as Make-A-Video and Video LDM. Compared to the other two methods, our approach can generate noticeable and accurate motion.
%For this assessment, we utilize DALL-E 2 and Midjourney to generate key frames, known for their superior aesthetic quality compared to Stable Diffusion.

\subsection{Ablation Studies}
We conduct ablation studies to validate our design choices concerning appearance injection and shifted noise training. For efficiency purposes, we adopt several different settings from the experiments used for system comparison. First, models employing different options are trained using a 1M subset of the filtered WebVid-10M dataset. Each model undergoes training for 64K steps (equivalent to one epoch) with a batch size set at 16. Second, during inference, we directly generate 17 frames without using the TSR module. Third, for the zero-shot FVD and IS evaluation on UCF101, we uniformly select 2K samples instead of using the entire 10K test set. It's notable that while using this smaller 2K-sample test set, the absolute FVD values are higher compared to those derived from the larger 10K-sample test set for the same model.

\subsubsection{Appearance Injection}
In an image\&text-to-video model, the most important design choice is how to inject the appearance information into the primary U-Net of the generation model. 

\noindent\textbf{Concatenation (Concat)}. A common approach in related work \cite{blattmann2023align,li2023videogen} is to direct concatenation of the latent features from the reference image to the noise input of the U-Net. 

\noindent\textbf{Addition to Decoder (Add-to-Dec)}. Our approach, however, adopts an AppearNet, akin to ControlNet for structure control. In the vanilla ControlNet, embeddings from the ControlNet are added to the decoder of the U-Net. We employ a similar operation in this setting.

\begin{figure}[t!]
  \centering
  \includegraphics[width=1.0\linewidth]{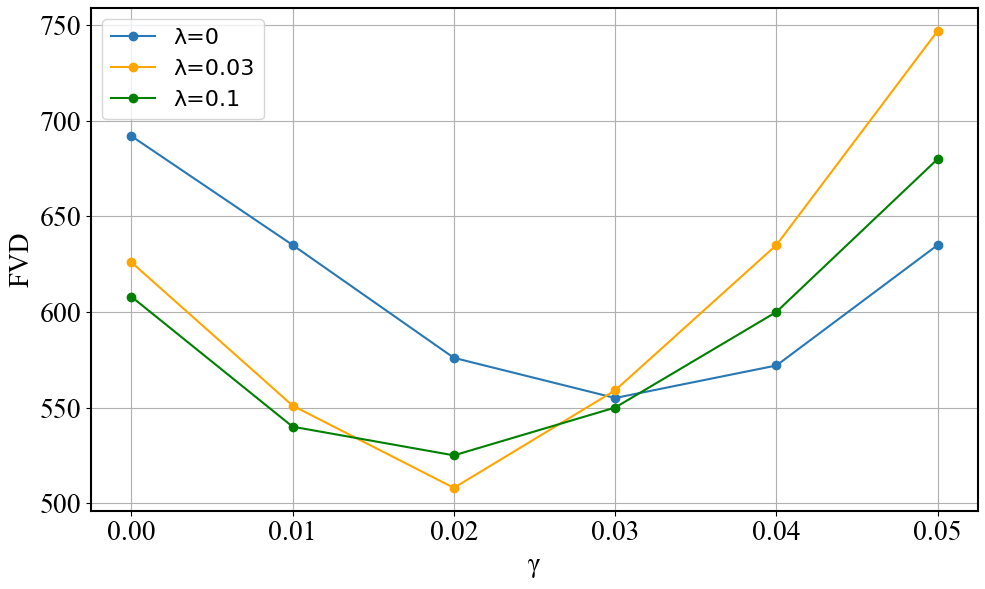}
    \vspace{-1.5em}
  \caption{UCF-101 Zero shot FVD across different \(\lambda\) and \(\gamma\).}
  \label{fig:abl_noise}
    \vspace{-1.5em}
\end{figure}

\begin{figure*}[ht]
  \centering
  \includegraphics[width=1.0\linewidth]{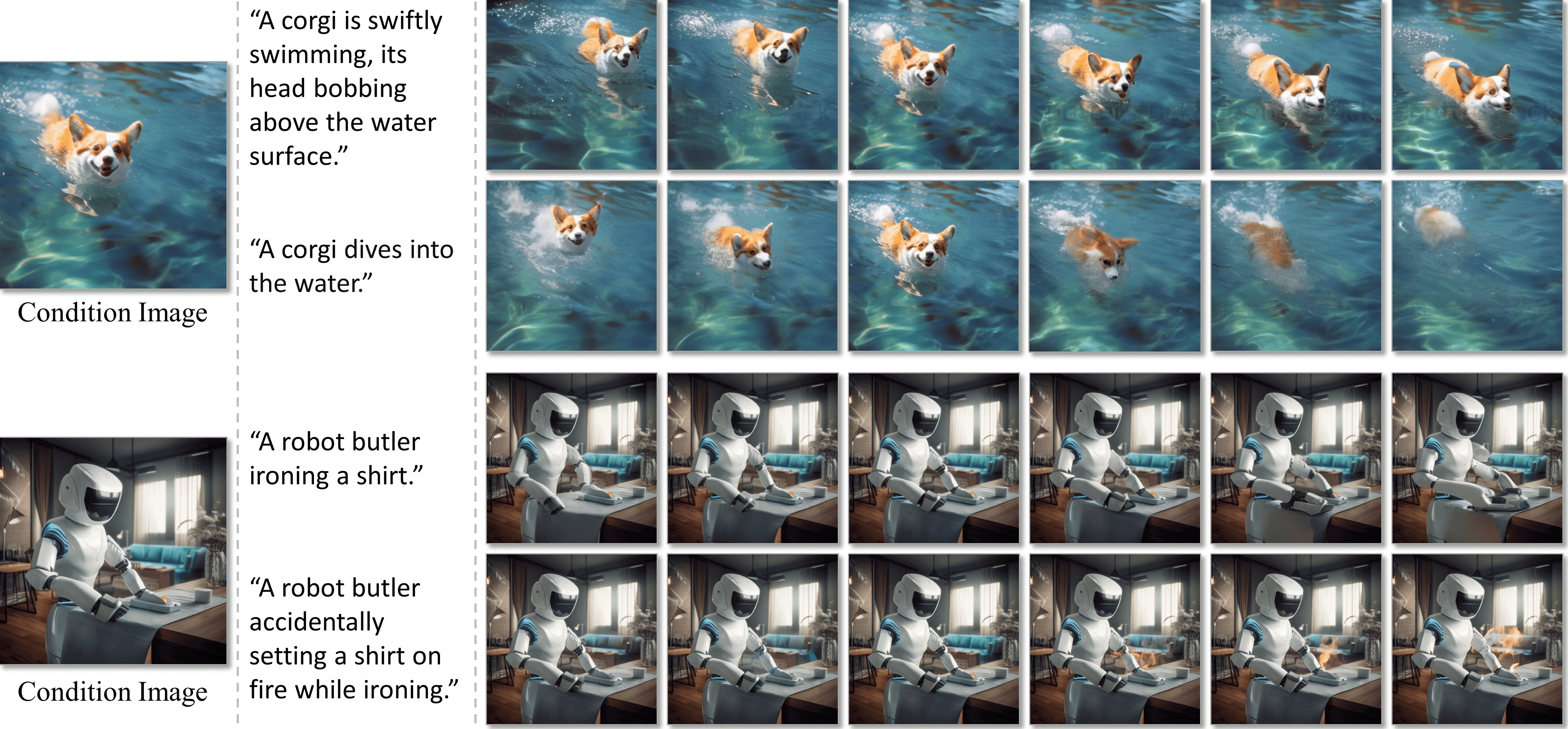}
   \vspace{-1.2em}
  \caption{We present the video generation results utilizing different prompts while conditioned on the same image. 
  %visualization, same condition image with difference prompt
  }
  \vspace{-1.2em}
  \label{fig:abaltion_text}
\end{figure*}

\noindent\textbf{Addition to Encoder and Decoder (Add-to-EncDec)}. Considering that the reference image contains more appearance details than the structural information in ControlNet, we propose injecting appearance into both the encoder and the decoder of the U-Net. This improvement is expected to elevate generation quality through a more comprehensive integration of appearance features.

\noindent\textbf{Addition to Encoder and Decoder with SPADE (Add-to-EncDec-SPADE)}. Expanding further, we integrate the SPADE technique, commonly used in image generation models, by infusing information into the GroupNorm layers of the U-Net. This final design constitutes the core of our method, MicroCinema.

\cref{tab:ucf_inject_ablation} presents a comparative analysis of the zero-shot FVD performance among these four design choices. The results clearly demonstrate that our final model achieves the most superior performance.

\begin{table}[t]
\small
    \centering
        \caption{\small Ablation study on UCF-101 for appearance injection methods.\vspace{-1em}}
    \label{tab:ucf_inject_ablation}
    
    % \resizebox{\linewidth}{!}{%
    \begin{tabular}{l  c c c}
        \toprule
        \textbf{Method} &  Zero-Shot & IS ($\uparrow$) & FVD ($\downarrow$)  \\
        \midrule
        Concat & Yes  & 15.83 & 688.92 \\
        Add-to-Dec  & Yes  & 27.90 & 589.59 \\
        Add-to-EncDec  & Yes  & 27.25 & 525.02 \\
        Add-to-EncDec-SPADE  & Yes & 29.63 & 508.56 \\
        \bottomrule
        \vspace{-2.2em}
    \end{tabular}
    % }
\end{table}
\vspace{-1em}
\subsubsection{Appearance Noise Prior}
Another key mechanism we propose for injecting appearance information into the image\&text-to-video generation network is the Appearance Noise Prior. One crucial and intricate parameter within this mechanism is the proportion, denoted by $\lambda$, determining the addition of the reference image to the noise input of the diffusion model. Selecting an optimal value for $\lambda$ involves balancing potential harm to the pre-trained image generation model and the advantages gained from additional information.

This set of ablation studies aims to empirically identify the most effective parameter for use with Appearance Noise Prior. Alongside $\lambda$, which we test at values of 0 (no Appearance Noise Prior), 0.03, and 0.1. Besides, according to our formulation, an appropriate amount of appearance may also help during the inference stage. Therefore, we also explore the impact of adding extra $\gamma \bm{z^c}$ to $ \epsilon$ during the inference stage. Therefore, the sampling noise during the inference stage is $(\lambda + \gamma) \bm{z^c} + \bm \epsilon_n$, where $\epsilon_n$ is sampled from $\mathcal{N}(\bm 0, \bm I)$. 

\cref{fig:abl_noise} shows the FVD scores across various combinations of $(\lambda, \gamma)$. We find that the lowest FVD score occurs when $\lambda=0.03$ and $\gamma=0.02$. Notably, this configuration leads to a substantial reduction in FVD compared to the baseline ($\lambda=0$, $\gamma=0$), dropping from 692 to 508, alongside a notable increase in IS from 18.5 to 29.6.

\subsection{Control in Image\&Text-to-Video Model}
Our image\&text-to-video model relies on both a reference image and a text prompt for conditioning. Our findings emphasize that the reference image's quality profoundly influences the resultant video quality. Consequently, both the text caption and the text-to-image model used to generate the reference image significantly impact the system's performance. We simplify our experiments by using the base image\&text-to-video model without using the temporal super-resolution component. In this setup, we adopt the resulting model to generate 17 frames with 10K samples on UCF101 for evaluating IS and FVD.

\cref{tab:ucf_prompt} illustrates the influence of various prompts on the model's generated outputs. We utilize the state-of-the-art SDXL model for text-to-image generation. Within the table, ``simple'' denotes a straightforward prompt created by connecting ``a video of'' with the motion tag, while ``LLaVA-1.5'' signifies a generated caption via the LLaVA-1.5 model \cite{liu2023improvedLLaVA,liu2023LLaVA} using the key frame as input. Results indicate that a well-crafted prompt correlates with higher-quality videos generated by the model.
\begin{table}[t]
\small
    \centering
        \caption{\small Evaluation on UCF-101 using different text prompts. SDXL is used as the first stage model.\vspace{-1em}}
    \label{tab:ucf_prompt}
    
    % \resizebox{\linewidth}{!}{%
    \begin{tabular}{l  c c c}
        \toprule
        \textbf{Method} &  Prompt & IS ($\uparrow$) & FVD ($\downarrow$)  \\
        \midrule
        MicroCinema & Simple  & 29.79 & 374.05 \\
        MicroCinema  & LLaVA-1.5  & 32.07 & 336.40 \\
        \bottomrule
        \vspace{-2.2em}
    \end{tabular}
    % }
\end{table}

Moreover, we assess the impact of employing different Text-to-Image (T2I) models. \cref{tab:ucf_1stage} underscores the substantial influence of T2I models on the FVD and IS of the generated videos. Notably, the design of MicroCinema affords us the flexibility to integrate various T2I models for generating the first-stage reference image, with potential performance enhancements stemming from advancements in text-to-image models.

\begin{table}[t]
\small
    \centering
        \caption{\small Evaluation on UCF-101 using different text-to-image models. Prompts generated by LLaVA-1.5 are utilized.\vspace{-1em}}
    \label{tab:ucf_1stage}
    
    % \resizebox{\linewidth}{!}{%
    \begin{tabular}{l  c c c}
        \toprule
        \textbf{Method} &  Frist Stage Model & IS ($\uparrow$) & FVD ($\downarrow$)  \\
        \midrule
        MicroCinema  & SD-2.1  & 31.25 &  412.53\\
        MicroCinema  & SDXL  & 32.07 & 336.40 \\
        \bottomrule
        \vspace{-2.2em}
    \end{tabular}
    % }
\end{table}

Lastly, we demonstrate the controllability of text prompts on the model's output in \cref{fig:abaltion_text}. Conditioning the image\&text-to-video model on the same image but varying text prompts results in significantly different videos aligned with their respective text prompts. This exemplifies the high level of control embedded within our model.
\section{Conclusion }
\label{sec:conclusion}
We presented MicroCinema, an innovative text-to-video generation approach that employs the Divide-and-Conquer paradigm to tackle two key challenges in video synthesis: appearance generation and motion modeling. Our strategy employs a two-stage pipeline, utilizing any existing text-to-image generator for initial image generation and subsequently introducing a dedicated image\&text-to-video framework designed to focus on motion modeling. To improve motion capture, we propose an Appearance Injection Network structure, complemented by an appearance-aware noise prior. Experimental results showcase MicroCinema's superiority, achieving a state-of-the-art zero-shot Frechet Video Distance (FVD) of \textbf{342.86} on UCF101 and \textbf{377.40} on MSR-VTT. We anticipate our research will inspire future advancements in this direction.

{
    \small
    \bibliographystyle{ieeenat_fullname}
    \bibliography{main}
}

\clearpage
%\setcounter{page}{1}
%\maketitlesupplementary

\appendix

\section{More Qualitative Results}
%We provide the generated results from diverse text prompts in video format in the folder named VideoResults.
\subsection{The Properties of Appearance Noise Prior.}
\begin{figure*}[ht]
  \centering
  \includegraphics[width=1.0\linewidth]{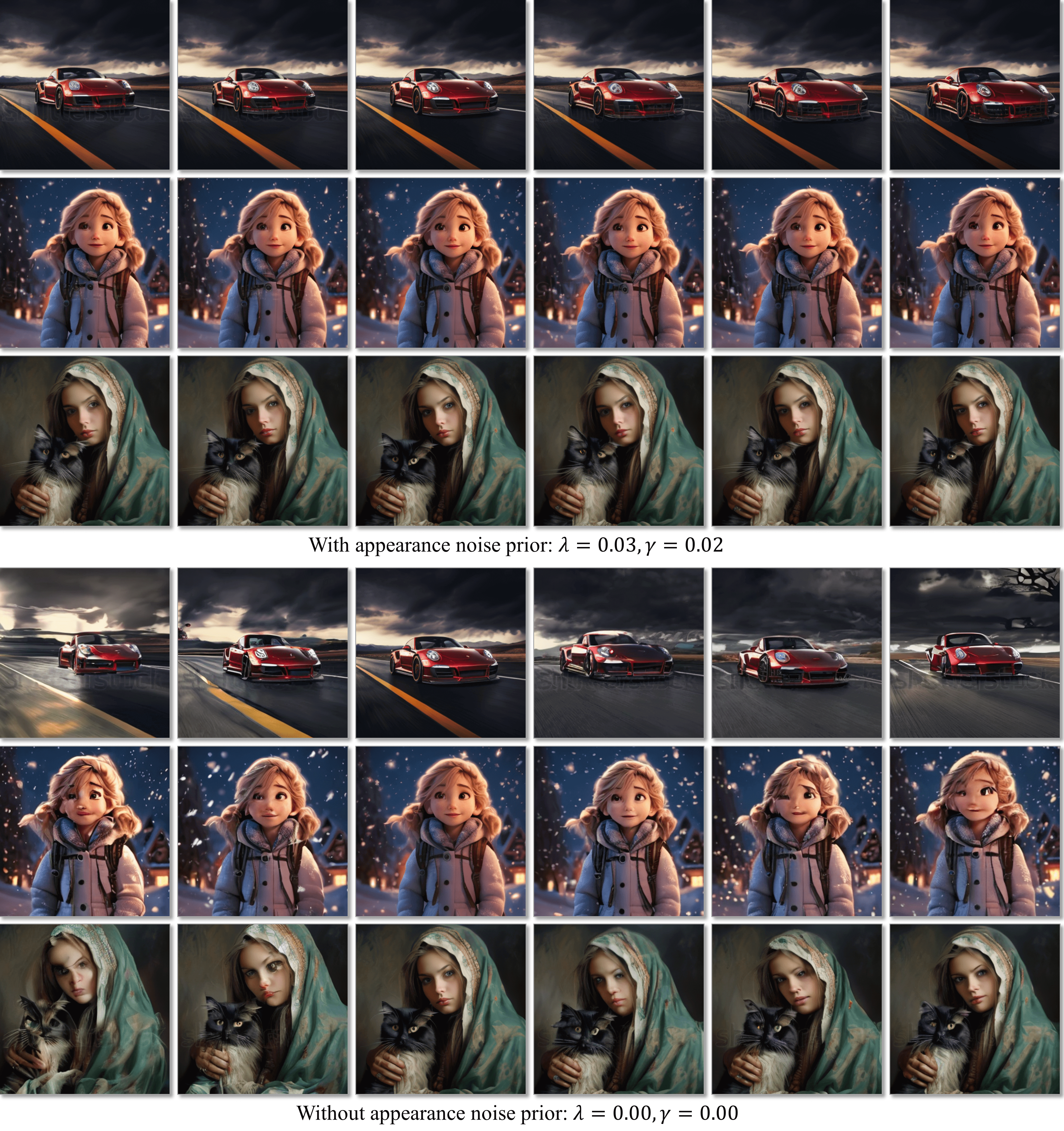}
   \vspace{-1.2em}
  \caption{The appearance noise prior is very useful for the model in maintaining the appearance of beautiful images. The prompts for the three videos, arranged from top to bottom, are as follows: 'Red Porsche running on the road, high resolution, 8k', 'Disney animation style, One frosty day, when snow blanketed everything like a white quilt, a little girl named Zosia was coming home from school. With gloves keeping her hands warm and a cozy jacket, she walked along the path', and 'Persian cat on a beautiful Polish woman.'}
  \vspace{-1.2em}
  \label{fig:vis_ablation_prior_320}
\end{figure*}

we fist examine the influence of the Appearance Noise Prior on the quality of generated video results by varying the parameters $\lambda$ and $\gamma$. As illustrated in \cref{fig:vis_ablation_prior_320}, videos generated with the integration of the Appearance Noise Prior display heightened coherence and superior quality in comparison to those generated without this prior. The introduced prior proves beneficial by endowing the model with enhanced capabilities to preserve distinctive characteristics of input images, even when they deviate from the training data in WebVid-10M.

\begin{figure*}[ht]
  \centering
  \includegraphics[width=1.0\linewidth]{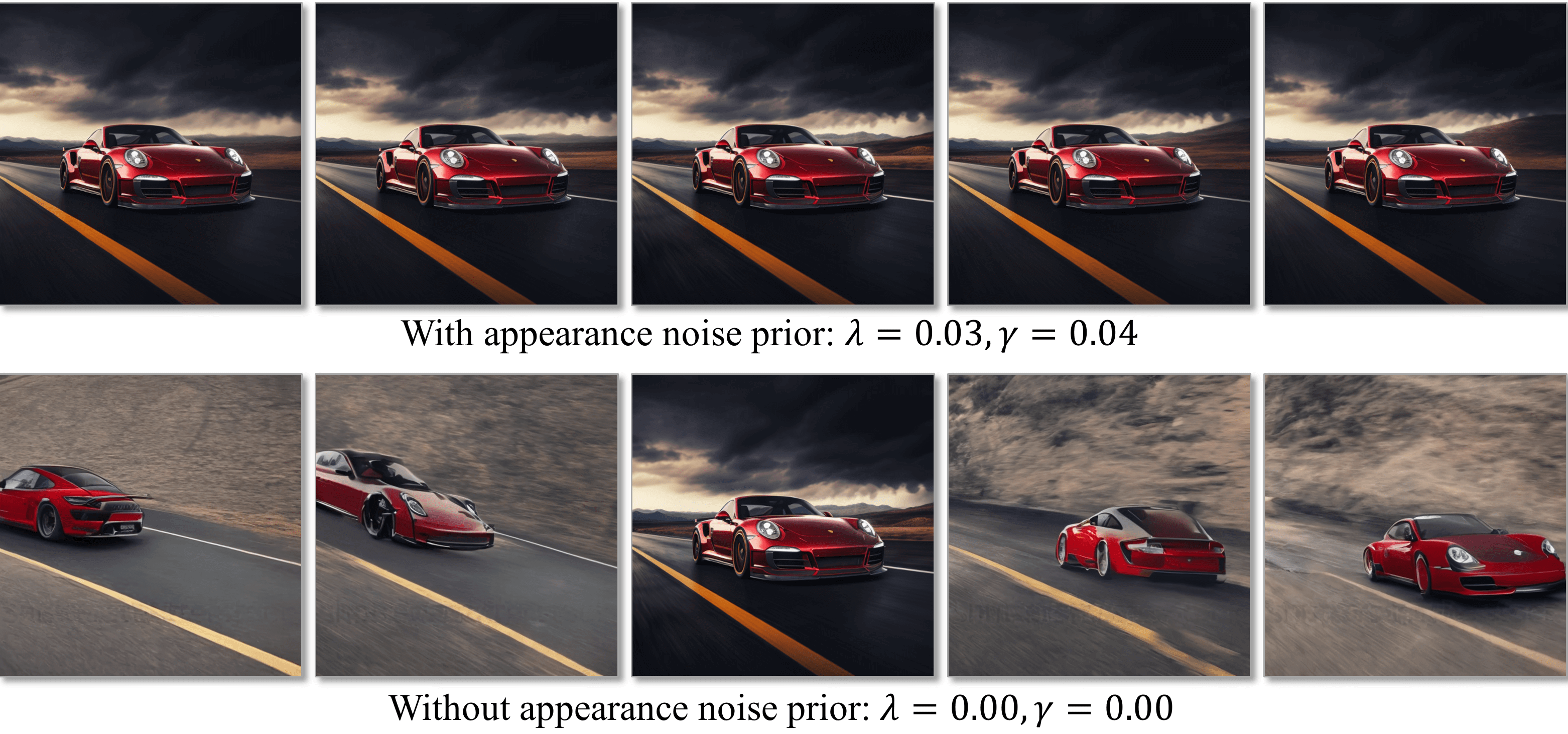}
   \vspace{-1.2em}
  \caption{The appearance noise prior enables the model to produce reasonable videos when the inference resolution (512x512) differs from the resolution (320x320) used during training.}
  \vspace{-1.2em}
  \label{fig:vis_ablation_prior_512}
\end{figure*}

Besides, our empirical findings indicate that adjusting the ratio of the Appearance Noise Prior contributes to the production of high-resolution videos by our model. As illustrated in \cref{fig:vis_ablation_prior_512}, the model demonstrates effective generation of 512x512 resolution videos, surpassing its original training resolution of 320x320, thanks to the integration of the Appearance Noise Prior.

\begin{figure*}[ht]
  \centering
  \includegraphics[width=1.0\linewidth]{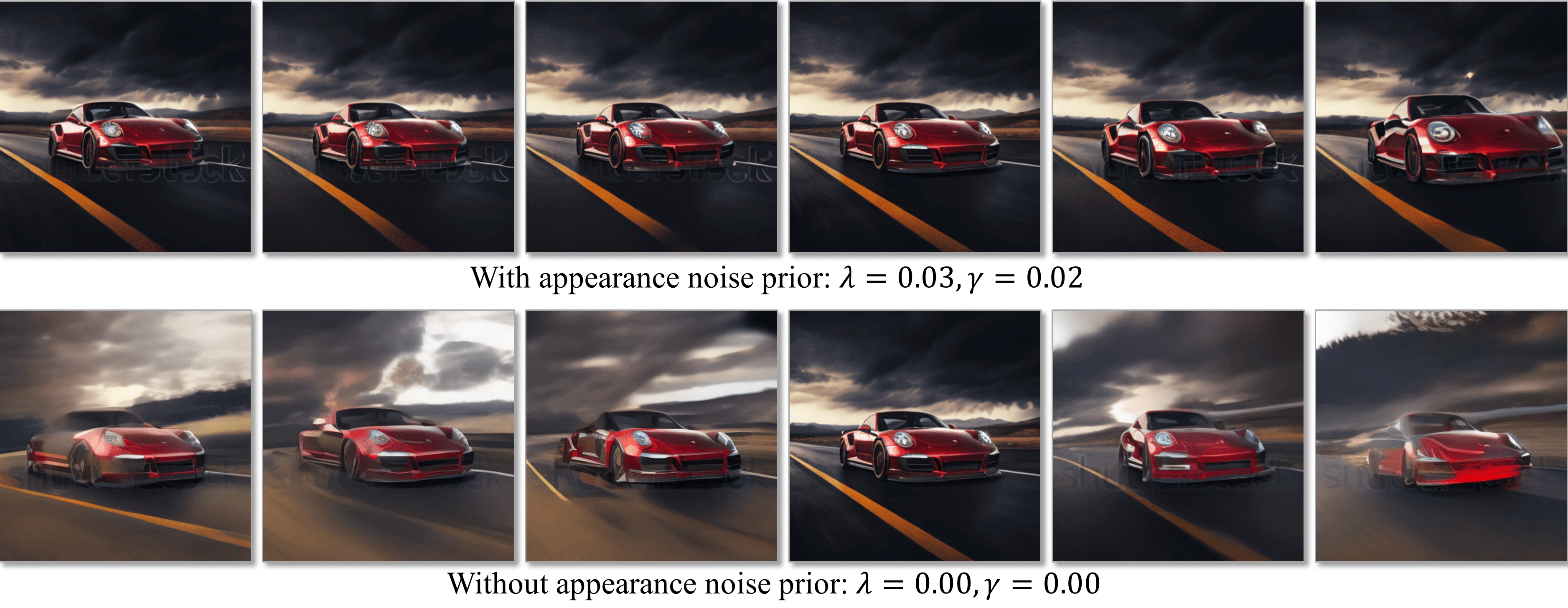}
   \vspace{-1.2em}
  \caption{The appearance noise prior enables the model to produce reasonable videos with sample steps of 5, especially for simple motion. This efficiency allows a video to be created in just 1.3 seconds.}
  \vspace{-1.2em}
  \label{fig:vis_ablation_prior_step5}
\end{figure*}

Additionally, we discover that the Appearance Noise Prior plays a crucial role in enhancing the efficiency of the diffusion process. As illustrated in \cref{fig:vis_ablation_prior_step5}, in situations involving simpler motion patterns, the integration of the Appearance Noise Prior empowers the network to produce satisfactory results even with a reduced sampling step count, set at 5. This decrease in steps significantly improves the efficiency of video production. For instance, employing our base image\&text-to-video model to generate a 9-frame video at 2 fps now requires only 1.3 seconds.

\subsection{More qualitative Results of MicroCinema}
\begin{figure*}[ht]
  \centering
  \includegraphics[width=1.0\linewidth]{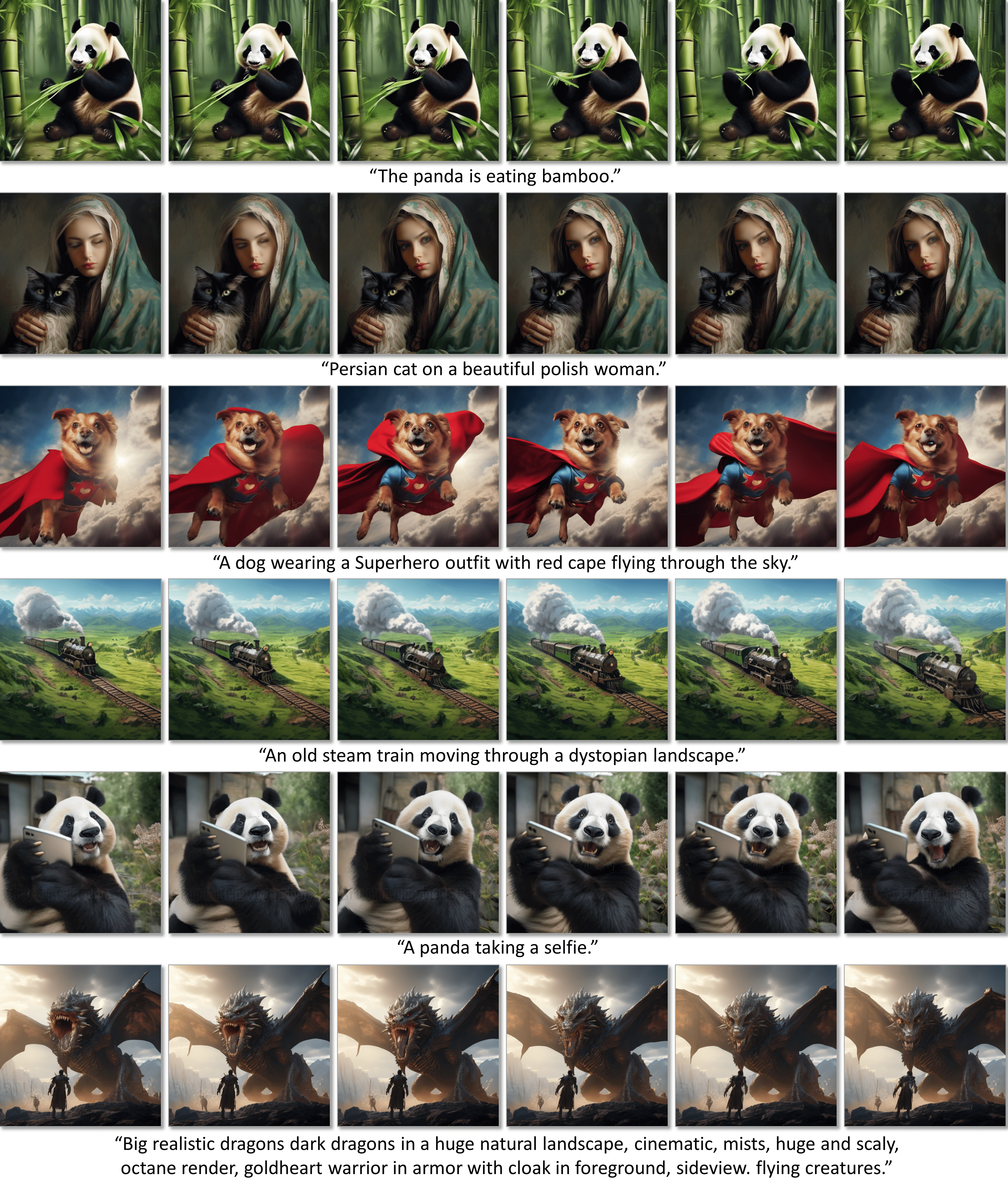}
   \vspace{-1.2em}
  \caption{Examples from MicroCinema demonstrate that our model is capable of generating exquisite imagery and crisp motion. Benefiting from a divide-and-conquer strategy, the model, though trained on the WebVid-10M dataset, can leverage given images to produce videos in various styles.
  %visualization, same condition image with difference prompt
  }
  \vspace{-1.2em}
  \label{fig:vis_result1}
\end{figure*}

\begin{figure*}[ht]
  \centering
  \includegraphics[width=1.0\linewidth]{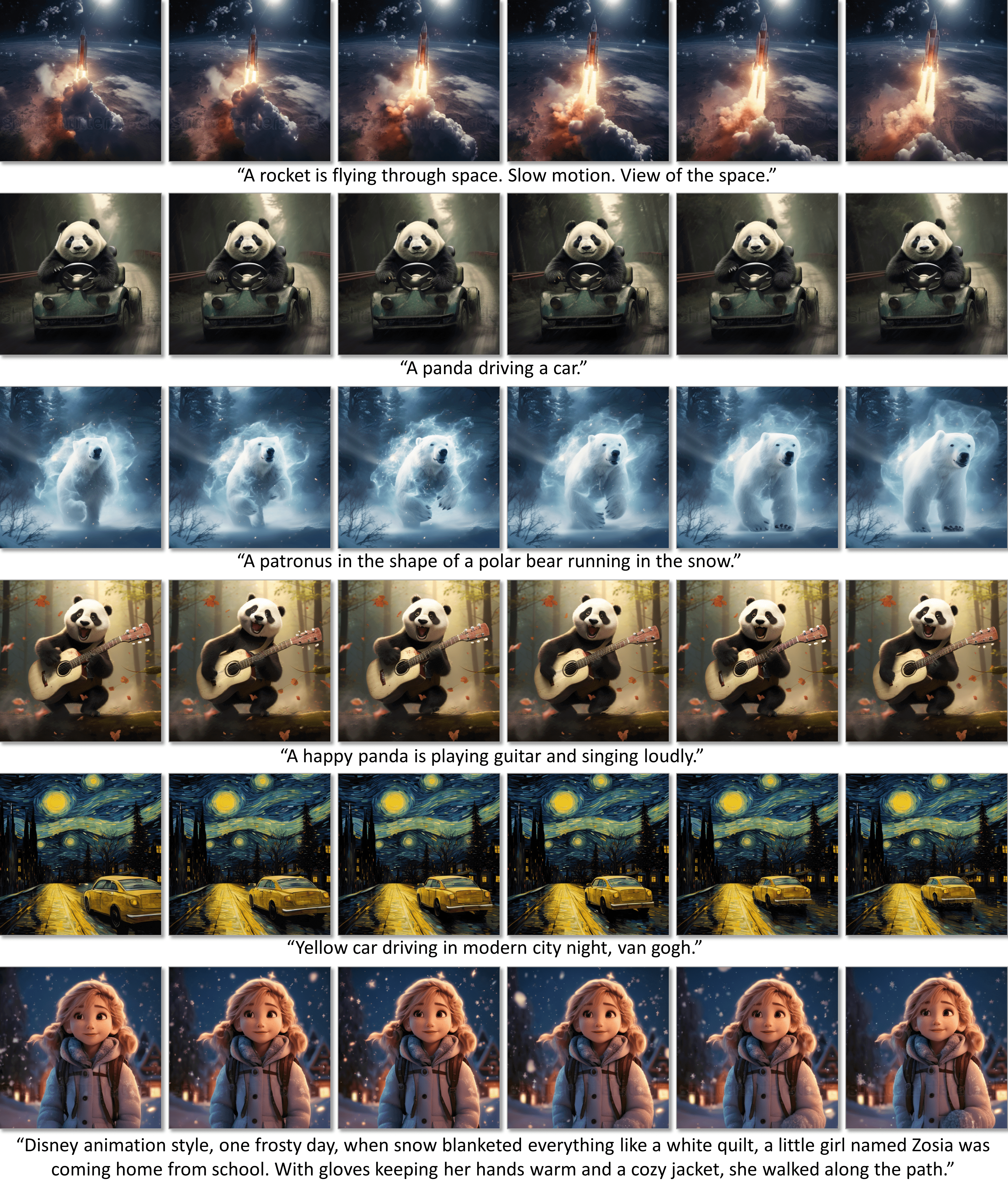}
   \vspace{-1.2em}
  \caption{More examples of MicroCinema.
  %visualization, same condition image with difference prompt
  }
  \vspace{-1.2em}
  \label{fig:vis_result2}
\end{figure*}

In this section, we present additional video generation results \cref{fig:vis_result1} and \cref{fig:vis_result2}. We utilize Midjourney as the initial stage text-to-video model.  It is evident that the videos generated through our method not only maintain aesthetic quality in imagery but also exhibit clear and coherent motion.

\subsection{Qualitative Comparison with Previous Work.}
\begin{figure*}[ht]
  \centering
  \includegraphics[width=1.0\linewidth]{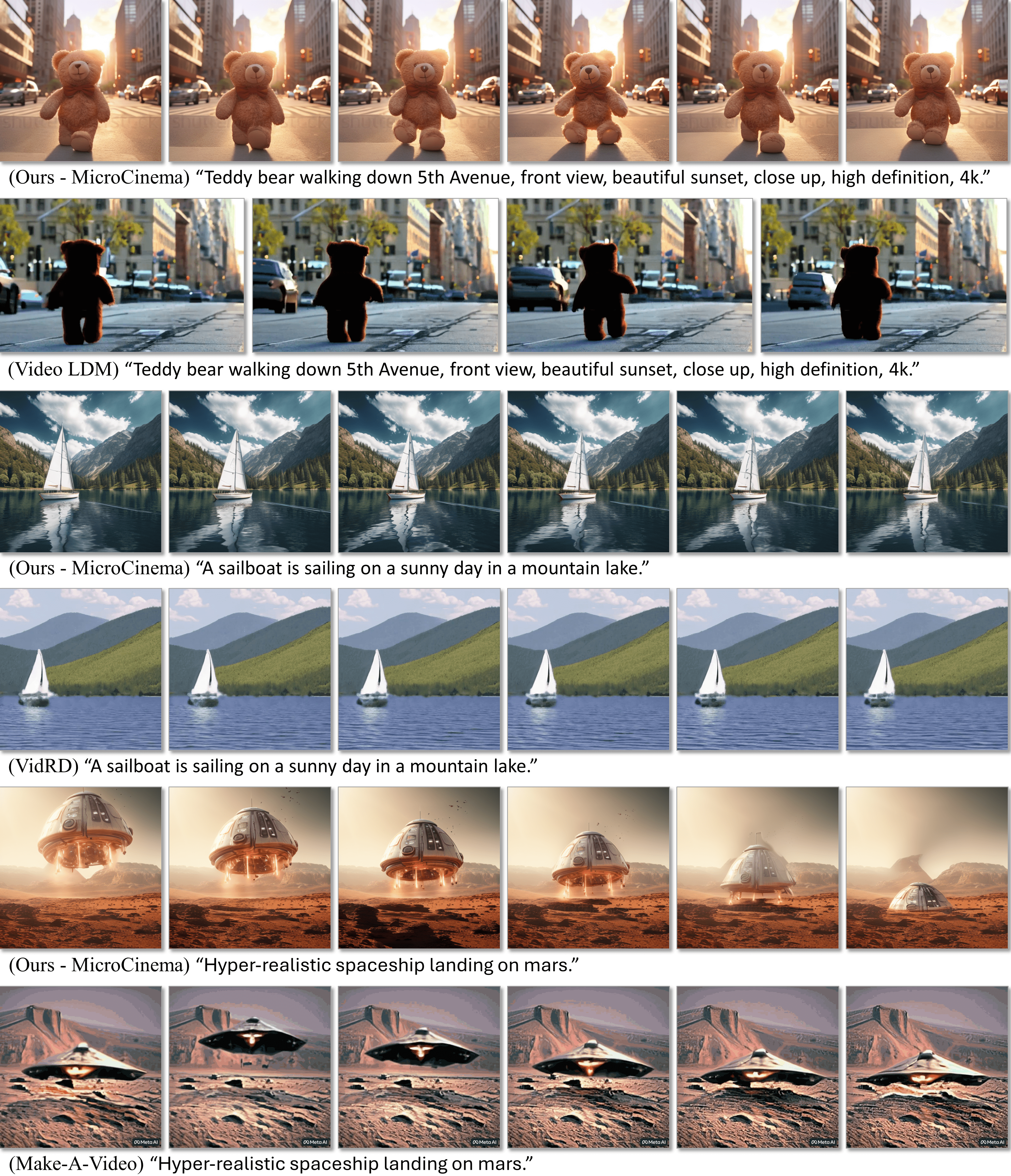}
   \vspace{-1.2em}
  \caption{We compare our method, MicroCinema, with Video LDM, VidRD, and Make-A-Video. Our videos exhibit clearer imagery and more distinct motion compared to Video LDM. In relation to VidRD, our image quality is significantly superior. When compared with Make-A-Video, our method demonstrates better image quality and text consistency.
  }
  \vspace{-1.2em}
  \label{fig:vis_compare0}
\end{figure*}

\begin{figure*}[ht]
  \centering
  \includegraphics[width=1.0\linewidth]{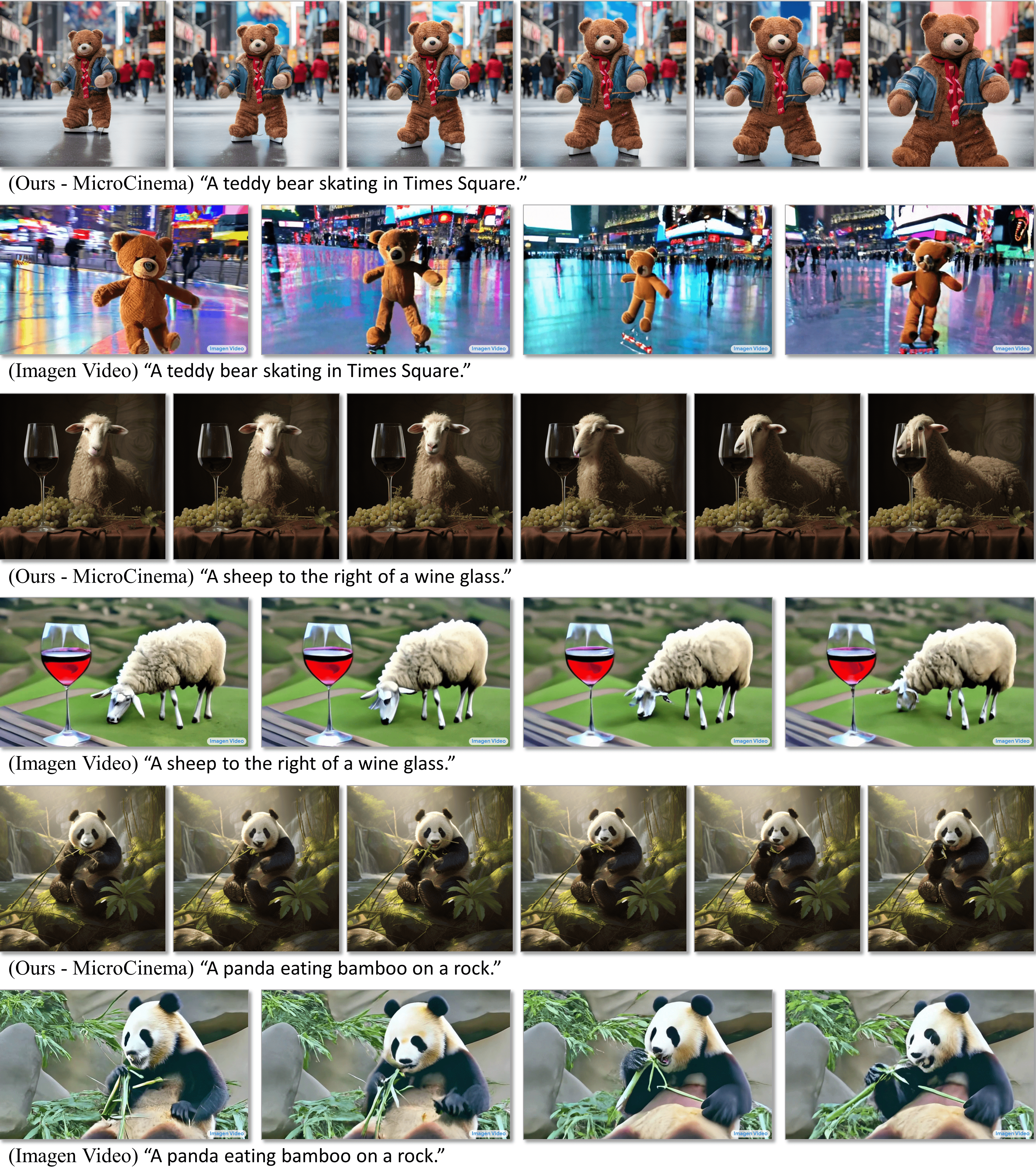}
   \vspace{-1.2em}
  \caption{We compare our method, MicroCinema, with Imagen Video. Our approach demonstrates superior temporal consistency, as evidenced (first row), compared to the Imagen Video method (second row). Furthermore, our method exhibits more exquisite image details, highlighted (third and fifth rows), while the videos produced by Imagen Video lack such detail (fourth and sixth rows).
  }
  \vspace{-1.2em}
  \label{fig:vis_compare1}
\end{figure*}

\begin{figure*}[ht]
  \centering
  \includegraphics[width=1.0\linewidth]{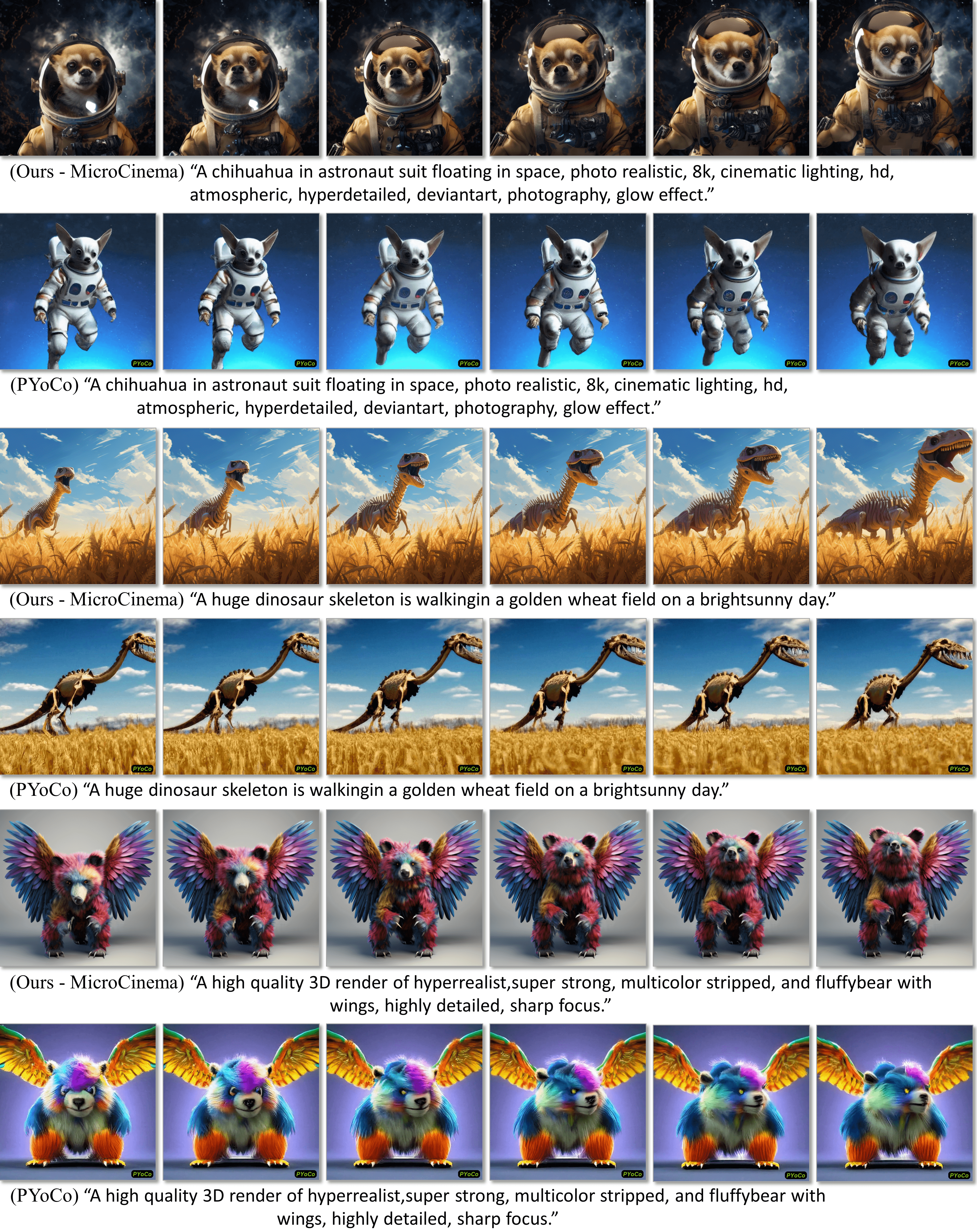}
   \vspace{-2em}
  \caption{We compare our method, MicroCinema, with PYoCo. Relative to PYoCo, MicroCinema excels in generating more intricate and visually appealing videos from complex descriptions. Our method showcases more pronounced motion and finer image details.
  %visualization, same condition image with difference prompt
  }
  \vspace{-1.2em}
  \label{fig:vis_compare2}
\end{figure*}

We provide additional examples for comparison with previous works in \cref{fig:vis_compare0}, \cref{fig:vis_compare1}, and \cref{fig:vis_compare2}. Our approach demonstrates the ability to generate visually stunning videos, akin to cinematic quality. In comparison to prior work, it showcases superior image quality, enhanced temporal consistency, greater stylistic diversity, and improved textual coherence.

\section{Proof of Appearance Noise Prior}
In this section, we present a proof of the compatibility of the Appearance Noise Prior with all ODE samplers. We demonstrate that incorporating the Appearance Noise Prior and employing new noise as supervision does not necessitate alterations to the sampler process itself. Instead, it only requires modifications to the initial noise during sampling.

\subsection{Denoising Diffusion Probabilistic Models}
Firstly, we introduce the standard framework of Denoising Diffusion Probabilistic Models (DDPM). The forward process in DDPM, when articulated in discrete form, is as follows:
\begin{equation}\label{eq:ddpm1}
\bm{z}_t = \sqrt{\bar{\alpha}_t} \bm{z}_0 + \sqrt{1 - \bar{\alpha}_t} \bm{\epsilon}, \quad t = 1, \ldots, T,
\end{equation}
\begin{equation}\label{eq:ddpm2}
\bm{z}_t = \sqrt{\alpha_t} \bm{z}_{t-1} + \sqrt{\beta_t} \bm{\epsilon}_{t-1}.
\end{equation}
The corresponding Stochastic Differential Equation (SDE) process of the DDPM can be represented by a unified expression, given by the following equation:
\begin{equation}\label{eq:ddpm_sde}
    \mathrm{d}\bm{z} = f(\bm{z}, x) \, \mathrm{d}x + g(x)\mathrm{d}\bm{w}, \quad x \in [0,1],
\end{equation}
where \(\bm{w}\) is a standard Wiener process. To derive the expressions for \( f(\bm{z}, x) \) and \( g(x) \), as \( T \) approaches infinity, two continuous functions, \( \bar{\alpha}(t) \) and \( \beta(t) \), can be defined:
\begin{equation}\label{eq:alpha_bar_t}
    \bar{\alpha}(x),x\in [0,1],\bar{\alpha}(x=\frac{t}{T})=\bar{\alpha}_t,
\end{equation}
\begin{equation}\label{eq:beta_t}
    \beta(x),x\in [0,1],\beta(x=\frac{t}{T})=T\beta_t,
\end{equation}
where \( \bar{\alpha}_t \) and \( \beta_t \) are the coefficients corresponding to those in equations \cref{eq:ddpm1} and \cref{eq:ddpm2}, respectively. By substituting \cref{eq:beta_t} into \cref{eq:ddpm2}, utilizing \( \alpha_t = 1 - \beta_t \), and considering the limit as \( T \to \infty, \beta_t \to 0 \), and subsequently applying a Taylor series expansion for approximation, equation \cref{eq:ddpm2} can be reformulated as follows:
\begin{equation}
\bm{z}_t = (1 - \beta(\frac{t}{T})\frac{1}{2T})\bm{z}_{t-1} + \sqrt{\frac{\beta(\frac{t}{T})}{T}} \, \bm{\epsilon}_{t-1}.
\end{equation}
By setting \( x= t/T \), and incorporating \( \bm{w} \) into the equation, we obtain:
\begin{equation}
\small
\bm{z}(x) - \bm{z}(x - \mathrm{d}x) = - \frac{\beta(x)}{2} \bm{z}(x - \mathrm{d}x)\mathrm{d}x + \sqrt{\beta(x)}\mathrm{d}\bm{w}.
\end{equation}
Upon simplification, we obtain the Stochastic Differential Equation (SDE) formulation of DDPM:
\begin{equation}
\mathrm{d}\bm{z}(x)= - \frac{\beta(x)}{2} \bm{z}(x)\mathrm{d}x + \sqrt{\beta(x)}\mathrm{d}\bm{w}, \quad x \in [0,1].
\end{equation}
Comparing with \cref{eq:ddpm_sde}, we can deduce:
\begin{equation}\label{eq:sde_fg}
    f(\bm{z}, x)=- \frac{\beta(x)}{2} \bm{z},\quad g(x)=\sqrt{\beta(x)}.
\end{equation}
For the SDE process described in \cref{eq:ddpm_sde}, the corresponding reverse Ordinary Differential Equation (ODE) process is represented by the following equation:
\begin{equation}
\mathrm{d}\bm{z} = f(\bm{z}, x) \, \mathrm{d}x - \frac{1}{2} g(x)^2 \nabla_{\bm{z}} \log p_x(\bm{z}) \, \mathrm{d}x.
\end{equation}
Given that \( \bm{z}(x) \) follows a Gaussian distribution \( N(\sqrt{\bar{\alpha}(x)}\bm{z}(0), (1-\bar{\alpha}(x))I) \), its score function can be related to the noise as follows:
\begin{equation}\label{eq:ddpm_score}
    \nabla_{\bm{z}} \log p_x(\bm{z})=-\frac{\bm{\epsilon}_{\bm{\theta}}(\bm{z}(x), x)}{\sqrt{1 - \bar{\alpha}(x)}},
\end{equation}
where \(\bm{\epsilon}_{\bm{\theta}}(\bm{z}(x), x)\) is estimated using the following loss function:
\begin{equation}
\mathcal{L}_{\bm{\theta}} =\mathbb{E}_{q(\bm{z}(x))} \left[ \left \| \bm{\epsilon}_{\bm{\theta}}(\bm{z}(x), x) - \bm{\epsilon} \right \|^2 \right],
\end{equation}
where \(q(\bm{z}(x))\) denotes the noisy data distribution of \(\bm{z}(x)\) and \( x \sim U[0,1] \). By utilizing equation \cref{eq:ddpm_score}, for DDPM models that implement the \(\epsilon\)-prediction, the reverse ODE process is articulated as follows:
\begin{equation}
\mathrm{d}\bm{z} = f(\bm{z}, x) \, \mathrm{d}x + \frac{\bm{\epsilon}_{\bm{\theta}}(\bm{z}(x), x)}{2\sqrt{1 - \bar{\alpha}(x)}} g(x)^2 \mathrm{d}x.
\end{equation}
By incorporating \cref{eq:sde_fg},  the final form can be derived as follows:
\begin{equation}\label{eq:ddpm_final}
\mathrm{d}\bm{z} = - \frac{\beta(x)}{2}\bm{z}\mathrm{d}x + \frac{\bm{\epsilon}_{\bm{\theta}}(\bm{z}(x), x)}{2\sqrt{1 - \bar{\alpha}(x)}} \beta(x) \mathrm{d}x.
\end{equation}

\subsection{Appearance Noise Prior}
To simplify notation, let \(\bm{\mu} =\lambda[\bm{z}^c, \bm{z}^c, ..., \bm{z}^c]\). Then the forward process of Appearance Noise Prior is change to:
\begin{equation}\label{eq:apn1}
\bm{z}_t = \sqrt{\bar{\alpha}_t} \bm{z}_0 + \sqrt{1 - \bar{\alpha}_t} (\bm{\epsilon}+\bm{\mu}), \quad t = 1, \ldots, T,
\end{equation}
\begin{equation}\label{eq:apn2}
\small
\bm{z}_t = \sqrt{\alpha_t} \bm{z}_{t-1} + (\sqrt{1 - \bar{\alpha}_t}-\sqrt{\alpha_t - \bar{\alpha}_t})\bm{\mu}+\sqrt{1 - \alpha_t}\bm{\epsilon}_{t-1}.
\end{equation}
Applying a transformation to the coefficient preceding \(\bm{\mu}\) in \cref{eq:apn2} yields:
\begin{equation}\label{eq:apn3}
\small
\bm{z}_t = \sqrt{\alpha_t} \bm{z}_{t-1} + \frac{1-\alpha_t}{\sqrt{1 - \bar{\alpha}_t}+\sqrt{\alpha_t - \bar{\alpha}_t}}\bm{\mu}+\sqrt{1 - \alpha_t}\bm{\epsilon}_{t-1}.
\end{equation}
Similar to equation (6), considering the limit conditions \( T \to \infty, \beta_t \to 0, \alpha_t \to 1 \), equation \cref{eq:apn3} can be reformulated as follows:
\begin{equation}
\small
\bm{z}_t = (1 - \beta(\frac{t}{T})\frac{1}{2T})\bm{z}_{t-1} + \frac{\beta(\frac{t}{T})}{2T\sqrt{1 - \bar{\alpha}_t}}\bm{\mu}+ \sqrt{\frac{\beta(\frac{t}{T})}{T}} \, \bm{\epsilon}_{t-1}.
\end{equation}
By setting \( x= t/T \) , and incorporating \( \bm{w} \) into the equation, we obtain:
\begin{equation}
\small
\mathrm{d}\bm{z}(x)= - \frac{\beta(x)}{2} \bm{z}(x)\mathrm{d}x + \frac{\beta(x)}{2\sqrt{1 - \bar{\alpha}(x)}}\bm{\mu}\mathrm{d}x+\sqrt{\beta(x)}\mathrm{d}\bm{w}.
\end{equation}
Comparing with \cref{eq:ddpm_sde}, we can deduce:
\begin{equation}\label{eq:anp_fg}
\small
    f(\bm{z}, x)=- \frac{\beta(x)}{2} \bm{z}+\frac{\beta(x)}{2\sqrt{1 - \bar{\alpha}(x)}}\bm{\mu}, \quad g(x)=\sqrt{\beta(x)}.
\end{equation}
Reverse ODE process can be represented by the following equation:
\begin{equation}\label{eq:anp_ode}
\mathrm{d}\bm{z} = f(\bm{z}, x) \, \mathrm{d}x - \frac{1}{2} g(x)^2 \nabla_{\bm{z}} \log p_x(\bm{z}) \, \mathrm{d}x.
\end{equation}
As we employ the following form of the loss function:
\begin{equation}
\mathcal{L}_{\bm{\theta}} =\mathbb{E}_{q(\bm{z}(x))} \left[ \left \| \bm{f}_{\bm{\theta}}(\bm{z}(x), x) - (\bm{\epsilon}+\bm{\mu}) \right \|^2 \right].
\end{equation}
Therefore, the relationship between the score function and the network's estimated value \( \bm{f}_{\bm{\theta}} \) becomes:
\begin{equation}\label{eq:anp_score}
    \nabla_{\bm{z}} \log p_x(\bm{z})=-\frac{\bm{f}_{\bm{\theta}}(\bm{z}(x), x)-\bm{\mu}}{\sqrt{1 - \bar{\alpha}(x)}}.
\end{equation}
By substituting \cref{eq:anp_score} and \cref{eq:anp_fg} into \cref{eq:anp_ode}, and noting that the coefficient preceding \(\bm{\mu}\) is eliminated, we obtain the final form of the Reverse ODE:
\begin{equation}\label{eq:anp_final}
\mathrm{d}\bm{z} = - \frac{\beta(x)}{2}\bm{z}\mathrm{d}x + \frac{\bm{f}_{\bm{\theta}}(\bm{z}(x), x)}{2\sqrt{1 - \bar{\alpha}(x)}} \beta(x) \mathrm{d}x.
\end{equation}
In the context of the Appearance Noise Prior, \( \bm{f}_{\bm{\theta}} \) functions as the network's output, paralleled by \( \bm{\epsilon}_{\bm{\theta}} \) in the DDPM framework. Notably, \cref{eq:anp_final} and \cref{eq:ddpm_final} exhibit identical forms. This similarity enables the straightforward integration of existing ODE sampling algorithms, with the only requisite modification being the adjustment of the initial sampling noise.

\subsection{Implementation of Appearance Noise Prior}
The implementation of Appearance Noise Prior in noise prediction models is straightforward. Traditionally, noise is added and trained using samples from a standard Gaussian distribution. With the Appearance Noise Prior, we modify this approach by superimposing an image prior \( \lambda[\bm{z}^c, \bm{z}^c, ..., \bm{z}^c] \) onto the original noise, creating a new noise term for noise addition and supervision. During inference with ODE samplers, the initial sampling noise should be changed from \( \mathcal{N}(\bm{0}, \bm{I}) \) to \( \mathcal{N}(\bm{\mu}, \bm{I}) \), where \( \bm{\mu} = \lambda [\bm{z}^c, \bm{z}^c, ..., \bm{z}^c] \). To achieve more consistent results, the strength of the prior can be appropriately enhanced by adjusting \( \bm{\mu} \) to \( (\lambda + \gamma)[\bm{z}^c, \bm{z}^c, ..., \bm{z}^c] \), thereby improving the consistency of the generated videos.

\begin{figure*}[!htb]
  \centering
  \includegraphics[width=1.0\linewidth]{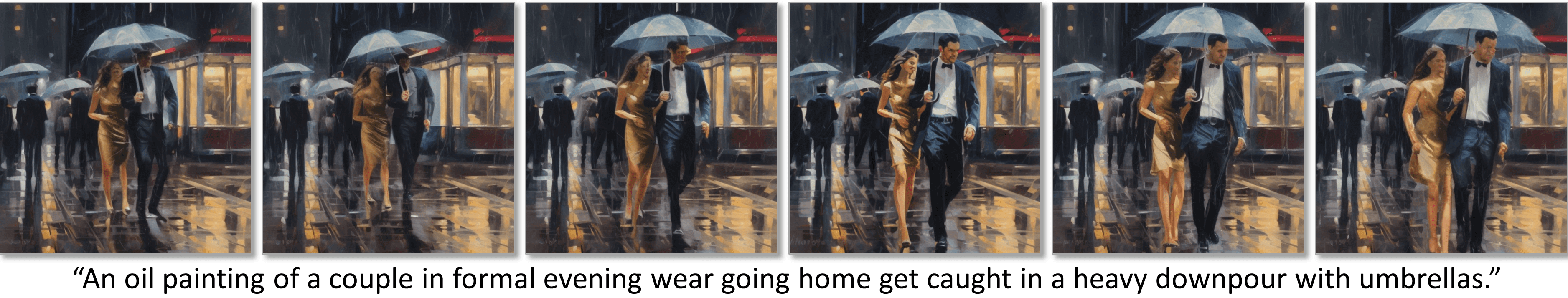}
   \vspace{-1.2em}
  \caption{When the facial features are extremely small, the model struggles to generate high-quality facial representations.}
  \vspace{-1.2em}
  \label{fig:vis_limitation}
\end{figure*}
\begin{figure*}[!htb]
  \centering
  \includegraphics[width=1.0\linewidth]{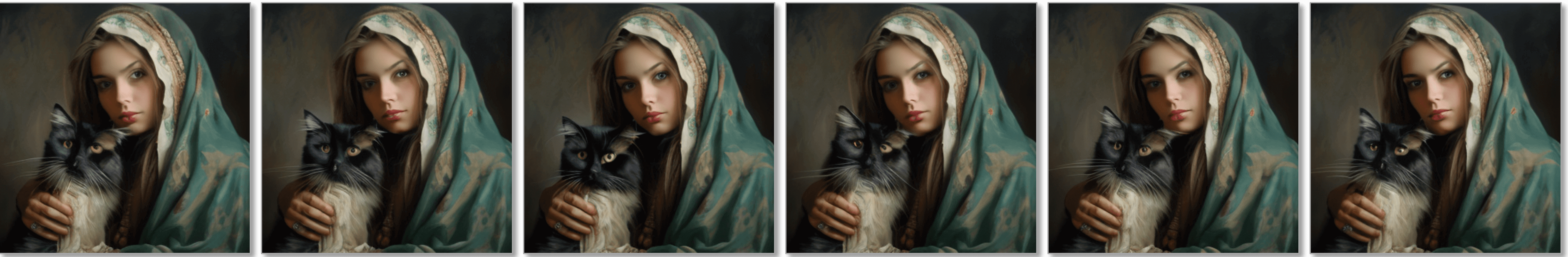}
   \vspace{-1.2em}
  \caption{When the facial features are extremely small,  the model performs significantly better.}
  \vspace{-1.2em}
  \label{fig:vis_limitation2}
\end{figure*}
\section{Implementation Details}
For text-to-image stage, we use SD2.1-Base and SDXL for Quantitative Experiments. Specific details of the samples are provided in the \cref{table:text2image}. And we use Midjourney and DALL-E 2 for Qualitative Results. In the \cref{table:hyperparameters_gen}, we present the specific details of our image\&text-to-video model. For the spatial layer, we utilized the SD2.1-Base model architecture and initial parameters. Additionally, we incorporated the VAE provided by SD2.1-Base, along with the CLIP text encoder, both of which were frozen during the training process. The \cref{table:parameters_num} displays the parameter count for each component; the image\&text-to-video model possesses 2.0 billion parameters, which were actively trained, while the spatial learning rate was set to one-tenth of the temporal learning rate.
\label{appendix:implementation_details}

\begin{table}[ht]
\setlength{\tabcolsep}{4pt}
\caption{Text-to-Image Model sampling parameters, generation time test on one A100-80GB.}
\centering
\resizebox{1.0\linewidth}{!}{
    \begin{tabular}{l|cc}
    \toprule
    \textbf{Sampling Parameters} &  \textbf{SD2.1-Base} &  \textbf{SDXL} \\
    \midrule
    Sampler & \multicolumn{2}{c}{EulerEDM} \\
    Steps & \multicolumn{2}{c}{50} \\
    Text guidance scale & \multicolumn{2}{c}{7.5}\\
    Image resolution & 512x512 & 1024x1024 \\
    Generation time & 2 s & 9 s \\ 
    \bottomrule
\end{tabular}%
}
\label{table:text2image}
\end{table}

\begin{table}[ht]
\setlength{\tabcolsep}{4pt}
\caption{Number of Model Parameters}
\centering
\resizebox{0.8\linewidth}{!}{
    \begin{tabular}{lc}
    \toprule
    \textbf{Model Name} &  \textbf{\# Params}  \\
    \midrule
    Base image\&text-to-video & 2 B \\
    CLIP text encoder  &  354 M \\
    VAE  &  84 M\\
    
    \bottomrule
\end{tabular}%
}
\label{table:parameters_num}
\end{table}

\section{Limitations}
Our method is based on the latent diffusion approach of SD2.1, utilizing an SD-pretrained VAE to encode images into the latent space. Currently, the VAE exhibits limited reconstruction capabilities for small objects, particularly small faces, leading to sub-optimal performance in these cases, as illustrated in the \cref{fig:vis_limitation}. Conversely, the model performs significantly better with larger faces, also demonstrated in the \cref{fig:vis_limitation2}. To address this issue, it is necessary to re-train the VAE with increased  channel size.

Another limitation of our approach is that we focused solely on temporal super-resolution (TSR) without incorporating spatial super-resolution (SSR). Ideally, a joint spatial-temporal super-resolution process could potentially achieve further improvements in the quality of the generated videos. This will be one of our future work. %This is partly due to the lower resolution nature of the WebVid-10M dataset. If trained on a higher-resolution dataset with both TSR and SSR, the model 

\begin{table*}[ht]
\setlength{\tabcolsep}{4pt}
\caption{Hyperparameters for our diffusion models are detailed as follows. In the spatial layers, we utilize the pretrained SD2.1-Base, as previously discussed. The term \({\gamma}^\dag\) represents a hyperparameter specific to the EulerEDM sampler, with \({\gamma}^\dag=0\) indicating the use of an ODE sampler.}
\centering
\resizebox*{0.9\textwidth}{!}{
    \begin{tabular}{lcc}
    \toprule
    \textbf{Hyperparameter} & \textbf{Base Image\&Text-to-Video Model} &  \textbf{Temporal Interpolation Model}  \\
    \midrule\midrule
    \textbf{Temporal Layers}  &  &\\
    \textit{Architecture}  &  &\\
    Input shape (C,N,H,W) & 4,9,40,40 & 4,5,40,40 \\
    Model channels & \multicolumn{2}{c}{320} \\
    Channel multipliers & \multicolumn{2}{c}{[1,2,4,4]}\\
    Attention resolutions  & \multicolumn{2}{c}{[4,2,1]}\\
    Head channels & \multicolumn{2}{c}{64}\\
    Positional encoding & \multicolumn{2}{c}{Sinusoidal}\\
    Temporal conv kernel size & \multicolumn{2}{c}{3,1,1}\\
    Temporal attention size & 9,1,1  & 5,1,1\\
    % \midrule
    % \emph{Image Conditioning} & &\\
    % $\dim c_{S}$  & 5 &5\\
    % Context channels & 128& 128\\
    \midrule
    \emph{Image Conditioning} & & \\
    Condition frame & \(\bm{z}^c\) & \(\bm{z}^1, \bm{z}^N\) \\
    Extending into video   & Repeat  & Interpolate\\
    \midrule
    \emph{Text Conditioning} & & \\
    Embedding dimension & 1024 & - \\
    CA resolutions  & [4, 2, 1]  & -\\
    CA sequence length & 77 & -  \\
    Drop rate & 0.1 & 1.0 \\
    \midrule
    \emph{Training}  &  & \\
    \# train steps  & 800K & 40K  \\
    Learning rate & $2\times 10^{-5}$ & $2\times 10^{-5}$  \\
    Batch size per GPU  & 4 & 4 \\
    \# GPUs  & 4 & 4  \\
    GPU-type  &A100-80GB & A100-80GB \\
    Training data FPS & 2 & 8, 30\\
    Prediction mode & \multicolumn{2}{c}{eps-pred} \\
    \midrule\midrule
    \textbf{Diffusion Setup} \\
    Diffusion steps &  \multicolumn{2}{c}{1000} \\
    Noise schedule & \multicolumn{2}{c}{Linear} \\
    $\beta_{0}$ & \multicolumn{2}{c}{0.00085} \\
    $\beta_{T}$ & \multicolumn{2}{c}{0.0120} \\
    Appearance noise prior $\lambda$ & \multicolumn{2}{c}{0.03} \\
    \midrule\midrule
    \textbf{Sampling Parameters}  &  & \\
    Sampler & \multicolumn{2}{c}{EulerEDM} \\
    Steps & \multicolumn{2}{c}{50} \\
    ${\gamma}^\dag$ & \multicolumn{2}{c}{0}\\
    Text guidance scale & 7.5 & 1.0\\
    Appearance noise prior $(\lambda + \gamma)$ & \multicolumn{2}{c}{0.03+0.02}\\
    Generation Time & 12 s & 7 s \\
    \bottomrule
\end{tabular}%
}
\label{table:hyperparameters_gen}
\vspace{-0.4cm}
\end{table*}




\end{document}